%% file: example_paper.tex
\documentclass{article}

\usepackage{microtype}
\usepackage{graphicx}
\usepackage{subcaption}
\usepackage{booktabs}
\usepackage{tcolorbox}
\usepackage{wrapfig}
\usepackage{hyperref}
\usepackage{multirow}
\usepackage{colortbl}
\usepackage{simpleicons}

\definecolor{brickred}{rgb}{0.6, 0, 0} 

\usepackage[preprint]{report}
\renewcommand{\headrulewidth}{0pt} 
\usepackage{fontawesome5}
\usepackage{amsmath}
\usepackage{amssymb}
\usepackage{mathtools}
\usepackage{amsthm}
\usepackage{enumitem}
\usepackage{wrapfig}
\usepackage{float}
\usepackage{tcolorbox}

\usepackage{xcolor}
\usepackage{tcolorbox} 
\definecolor{acadred}{RGB}{160, 20, 20} 
\definecolor{acadredbg}{RGB}{250, 240, 240} 
\definecolor{hfYellow}{HTML}{FFD21E}
\definecolor{lightyellow}{RGB}{254, 249, 243}
\definecolor{yellow}{RGB}{224, 190, 120}
\definecolor{depthblue}{RGB}{92, 170, 255}
\definecolor{acadblue}{RGB}{0, 0, 123}

\definecolor{acadivorybg}{RGB}{251, 206, 177}
\hypersetup{
    colorlinks=true,
    linkcolor=acadblue,  
    citecolor=acadblue,  
    urlcolor=acadblue    
}

\newtcolorbox{researchquestion}{
    colback=lightyellow,       
    colframe=yellow,        
    boxrule=0.8pt,           
    arc=2pt,                 
    width=\linewidth,        
    left=4pt, right=4pt,     
    top=4pt, bottom=4pt,     
    before skip=24pt,         
    after skip=24pt,          
    fontupper=\sffamily\bfseries\centering 
}

\newtcolorbox{findings}{
    colback=lightyellow,       
    colframe=yellow,        
    boxrule=0.8pt,           
    arc=2pt,                 
    width=\linewidth,        
    left=4pt, right=4pt,     
    top=4pt, bottom=4pt,     
    before skip=8pt,         
    after skip=8pt,          
    fontupper=\centering 
}

\newtcolorbox{insightbox}[1]{
    colback=white,          
    colframe=acadred,       
    colbacktitle=lightyellow, 
    coltitle=acadred,       
    fonttitle=\bfseries,    
    title={#1},             
    boxrule=0.6pt,
    arc=2pt,
    attach boxed title to top left={yshift=-2mm, xshift=2mm}, 
    enhanced,               
    left=4pt, right=4pt, top=8pt, bottom=4pt
}

\newtcolorbox{definitionbox}{
    blanker,                
    borderline west={2pt}{0pt}{acadred}, 
    colback=acadredbg,      
    left=8pt, right=8pt, top=6pt, bottom=6pt,
    arc=0pt
}

\DeclareCaptionFont{acadredfont}{\bfseries}
\usepackage{sectsty} 

\usepackage{sectsty}
\sectionfont{\sffamily\bfseries} 
\subsectionfont{\sffamily\bfseries}

\usepackage[capitalize,noabbrev]{cleveref}
\usepackage{lmodern}
\usepackage{roboto}

\theoremstyle{plain}

\theoremstyle{definition}

\theoremstyle{remark}

\usepackage[textsize=tiny]{todonotes}

\icmltitlerunning{}

\makeatletter
\renewcommand{\paragraph}{\@startsection{paragraph}{4}{\z@}
    {-0.2ex plus -0.5ex minus -1ex} 
    {-1em}                            
    {\sffamily\bfseries}}
\makeatother

\begin{document}

\makeatletter
\icml@noticeprintedtrue 
\makeatother

\begin{tcolorbox}[
    colback=lightyellow,
    colframe=lightyellow,
    arc=4mm,
    boxsep=3mm, 
    width=\linewidth
]
    \vspace{3mm}
    \begin{center}
        {
             \sffamily \bfseries \fontsize{17}{20} \selectfont \raisebox{-0.18\height}{
        \includegraphics[height=1.4em]{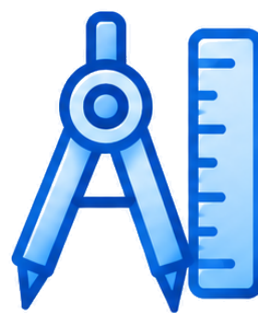}
    } \textsc{DepthBench}: Measuring How Residual Connections Enable More Computational Depth
        }
        \par
        \vspace{4mm}
        {
            \sffamily 
            \fontsize{10}{3}
            \selectfont
             \textbf{Keyu Wang}$^{1, 2, 3*}$, 
             \textbf{Yangyi Huang}$^{4*}$, 
             \textbf{Jiale Kang}$^{4}$,
             \textbf{David González-Martínez}$^{1, 2, 3}$, \\
             \textbf{Weiyang Liu}$^{4}$, 
             \textbf{Shiwei Liu}$^{1, 2, 3}$
        }
        \vspace{2mm} \\
        {
            \fontsize{10}{14}
            \selectfont
            \color{black!80}
            $^1$ELLIS Institute Tübingen, $^2$Max Planck Institute for Intelligent Systems,\\
            $^3$Tübingen AI Center, 
            $^4$The Chinese University of Hong Kong \\
            $^*$Equal contribution. 
            Correspondence to: {keyu.wang@tuebingen.mpg.de, sliu@tue.ellis.eu}
        }
    \end{center}
    \vspace{3mm}
    Depth is a natural way to increase the computational capacity in Transformers, yet the contribution of deeper layers can diminish as depth grows larger. Recent approaches enhance normalization (\text{e.g.}, LayerNorm Scaling) or residual connections (\text{e.g.}, mHC, AttnRes) to enable better information flow and depth utilization. However, it remains unclear whether they truly translate increased architectural depth into effective computational depth, and whether their reported gains stem from better access to information across depth, or unaccounted-for confounding factors. In this paper, we introduce \textbf{DepthBench}, a controlled benchmark for studying computational depth across various architectures. We systematically vary the width--depth aspect ratio ($d_{\text{model}}/n_{\text{layer}}$) from shallow--wide to deep--narrow shapes, while keeping the model size and pre-training recipe fixed. Across 10 representative architectures, we find that the benefit of allocating more capacity to depth is strongly architecture-dependent. Standard Pre-LN and most of its norm- and scaling-based variants provide little benefit and can even degrade performance as models become deeper and narrower, whereas HC and Full AttnRes improve consistently even at extreme deep shapes.  These gains extend beyond pre-training loss and consistently translate into improved domain-specific performance and effective computation. Controlled layer-level analyses further show that the gains of HC and Full AttnRes are associated with more effective utilization of additional layers, revealing distinct mechanisms of computational depth across architectures.  Overall, our results identify residual connection design as a key determinant of whether depth can serve as a meaningful scaling axis by enabling additional architectural depth to translate into effective computation.
    \noindent
\vspace{0.2cm}

\begin{center}
    \textbf{ \sffamily Code:} \href{https://github.com/keyu-wang-2002/DepthBench/tree/main}{\texttt{DepthBench}} 
  \quad 
  \textbf{  \sffamily Checkpoints:} \href{https://huggingface.co/aspect-ratio-scaling}{\texttt{HuggingFace}} 
\end{center}

\end{tcolorbox}

\makeatletter
\global\icml@noticeprintedtrue 
\makeatother

\begin{figure}[H]
    \centering
    \includegraphics[width=1.0\linewidth]{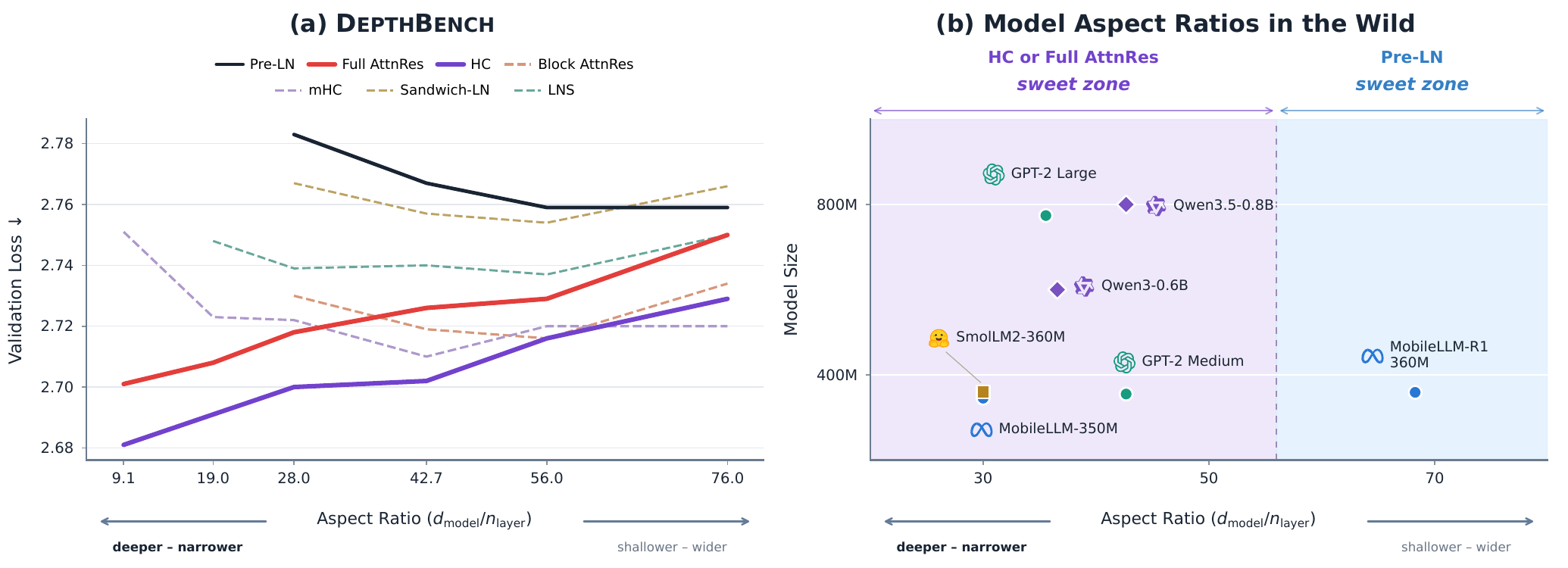}
    \caption{{\textbf{Left:} Validation loss across  aspect ratios under an approximately fixed 400M parameter budget and shared pre-training recipe. \textbf{Right: } Model sizes and aspect ratios of representative industry-released models with Pre-LN architecture, overlaid with shaded regions illustrating the contrasting shape preferences observed in \textsc{DepthBench}.}}
    \label{fig:placeholder}
\end{figure}






\section{Introduction}

Scaling depth is a fundamental way to increase the computational capacity of Transformers,  as more layers allow representations to undergo a longer sequence of nonlinear transformations \citep{depthlimits, logdepth}.  However, current LLMs suffer from the curse of depth \citep{lns}, where purely increasing architectural depth does not necessarily yield a proportional increase in model capacity. In conventional Pre-LN Transformers, residual connections make deep networks easier to optimize \citep{pre-ln}, but they do not ensure that every layer contributes useful computation \citep{lns}. As models become deeper, later-layer updates can become increasingly weak, redundant, or similar to those of preceding layers \citep{mix-ln, lns, attnres, inversedepthscaling}. Consequently, additional layers may therefore increase nominal depth without comparable gains in effective computation or model quality.

Recent work addresses this problem by modifying normalization or residual propagation. For instance, LayerNorm Scaling (LNS) \citep{lns} and KEEL \citep{keel} refine normalization,  whereas AttnRes \citep{attnres}, HC \citep{hc}, and mHC \citep{mhc} redesign the residual pathway.
These methods report improved optimization and model quality at large scales, and related designs are being adopted in frontier LLMs such as Kimi K3 \citep{k3}  and DeepSeek V4 \citep{v4}.

However, it remains unclear whether they truly translate increased architectural depth into effective computational depth\footnote{Effective computational depth is the amount of sequential computation that a model can exploit through successive layers, as measured by the incremental benefit of additional layers or iterations under a controlled compute budget.}. Existing methods are typically evaluated with different architectures, training recipes, and system configurations. Their gains therefore cannot be cleanly attributed to improved optimization, better access to information across depth, preservation of diverse features, or differences in capacity and computational overhead.

We therefore still lack a unified and comprehensive understanding of which architectural designs can reliably translate more architectural depth into useful computational depth.  This motivates our primary research question: 
\begin{center}
\begin{tcolorbox}[width=0.9\textwidth]
\centering
\textit{Which architectural designs make additional depth computationally effective?}
\end{tcolorbox}
\end{center}

Answering this question determines when depth can serve as a reliable scaling axis rather than merely increasing the number of layers.

To address this question, we develop \textsc{DepthBench} to study when depth pays off under a fixed parameter budget. Varying the aspect ratio ($d_{\text{model}}/n_{\text{layer}}$) reallocates parameters between width and depth, enabling a direct comparison between shallow-wide and deep-narrow architectures with a broad range of aspect ratios for 10 representative residual designs while holding both the parameter count and training recipe fixed. We sweep multiple learning rates for every aspect ratio and report each configuration at its best-performing learning rate, ensuring that differences reflect the architecture rather than a suboptimal learning rate. We then rank model performance across configurations and conduct controlled layer-level analyses to determine whether each approach enables effective aspect ratio scaling by making better use of the greater depth.

Our results reveal several insights about how residual connections affect computational depth:

\begin{itemize}[leftmargin=*] 
    \item \textbf{Conventional residual connections do not benefit consistently from deeper architectures at iso-parameter (Section \ref{sec:main-results}).} 
    Pre-LN and its normalization/residual-scaling variants do not benefit from deeper, narrower architectures. Their optima remain at relatively large aspect ratios $42.7$--$76.0$ (e.g., $d_{\text{model}}=1120$ and $n_{\text{layer}}=20$), obscuring aspect ratio as a useful scaling axis.
    \item \textbf{New residual designs unleash the power of deeper models even with extremely small aspect ratios (Section \ref{sec:main-results}).} 
    HC and Full AttnRes continue to reduce pre-training loss as models become deeper-narrower under iso-parameter scaling, even at an extreme aspect ratio of 9.1 with $d_{\text{model}}=640$ and $n_{\text{layer}}=70$.  Effective residual design  thus turns aspect ratio into a practical scaling dimension.
    \item \textbf{Alternative residual mechanisms fundamentally change how computation evolves across depth (Section \ref{sec:depth-metric} and Section \ref{sec:depth-processing}).} 
    Full AttnRes and HC preserve heterogeneous, layer-specific transformations across layers, unlike the increasingly homogeneous deep-layer representations in Pre-LN.
    \item \textbf{The benefits of depth scaling extend beyond pre-training loss (Section \ref{sec:main-results} and Section \ref{sec:depth-metric}).}
    The gains of HC and Full AttnRes transfer to domain-specific evaluations. Layer-wise diagnostics further indicate that these methods make more effective use of the additional computational depth.
    \item \textbf{No free lunch: deep models introduce a systems-level efficiency trade-off (Section \ref{sec:efficiency-trade-off}).}
    Although deeper architectures can improve modeling performance, they increase compute and memory overhead while reducing hardware utilization. Realizing their benefits at scale therefore requires better infrastructure and kernel optimization.
\end{itemize}

\section{Preliminaries \& Setup} 
\label{sec:pre}

\subsection{Modern Deep Transformer Architectures} \label{sec:architectures}
\paragraph{Pre-LN Transformers with Residual Connections.}
Modern LLMs predominantly adopt the Pre-LN Transformer architectures \citep{pre-ln}, where layer normalization is applied before each transformation branch and the resulting update is added to the residual stream \citep{pre-ln-apply1, pre-ln-apply2}. Formally, a Pre-LN sublayer updates the hidden state as:
\begin{equation}
    h_{\ell}
    =
    h_{\ell-1}
    +
    F_{\ell}\!\left(
        \mathrm{LN}(h_{\ell-1})
    \right)
\end{equation}
where $F_{\ell}$ denotes the attention or feed-forward transformation at layer $\ell$, and $\mathrm{LN}$ is typically instantiated as RMSNorm \citep{rmsnorm}. Compared with Post-Layer Normalization (Post-LN) \cite{post-ln}, Pre-LN substantially improves training stability at large depth by preserving a direct residual pathway for gradient propagation \cite{pre-ln, mix-ln}.

\paragraph{Pre-LN Issues and Improvements.} 
While Pre-LN largely resolves the \textit{trainability} issue of deep Transformers \cite{pre-ln}, it does not necessarily guarantee \textit{effective compuational depth}: a model can be made very deep without fully utilizing its later layers \cite{mix-ln, lns}. This limitation has been widely described as the \textbf{Curse of Depth} \cite{lns} and \textbf{Pre-LN dilution} \cite{attnres}, where the contribution of deeper layers progressively weakens.

A growing line of architectures therefore modifies normalization and residual connections, seeking better depth utilization \cite{keel, lns, attnres, siamesenorm}. To provide a thorough evaluation, we categorize the architectures considered in this work into four groups, summarized in Table \ref{architecture_sketch}: 

\input{tables/architecture_sketch}

\begin{itemize}
\item \textbf{Baseline.}
We employ standard \textbf{Pre-LN} \cite{pre-ln} as our primary baseline.

\item \textbf{Normalization Variants.}
This category primarily modifies layer normalization.
We include \textbf{Sandwich-LN} (also called Peri-LN) \cite{sandwich-ln1, sandwich-ln2}, which introduces  normalization before and after transformation branch; \textbf{LNS} \cite{lns}, which adjusts the scale of LayerNorm outputs in Pre-LN; \textbf{DeepNorm} \cite{deepnorm} and \textbf{KEEL} \cite{keel}, which improve upon the Post-LN formulation to enable more stable gradient propagation in deep Transformers.

\item \textbf{Multi-stream Residuals.}
These architectures extend the conventional single residual stream into multiple streams.
We include \textbf{HC} \cite{hc} and \textbf{mHC} \cite{mhc}, which maintain and mix multiple residual streams to provide more flexible cross-layer information propagation.

\item \textbf{Cross-layer Access.}
Finally, this category enables each layer to directly access representations from earlier depths, rather than receiving information solely through recursive propagation from the immediately preceding layer.
We include \textbf{AttnRes} \cite{attnres}, which aggregates preceding hidden states through attention-based residual connections, and \textbf{MoDA} \cite{moda}, which dynamically combines KV cache representations from earlier layers.

\end{itemize}

\subsection{DepthBench Design}

\label{sec:aspect_ratio}

\paragraph{Controlled Width--Depth Aspect Ratio Scaling.}
Simply increasing depth at fixed width also increases model size and compute, confounding the effect of depth with generic scale.
\textsc{DepthBench} instead treats depth as a \emph{capacity allocation} choice, varying $d_{\mathrm{model}}/n_{\mathrm{layer}}$ from shallow--wide to deep--narrow configurations under an approximately fixed parameter budget.


\paragraph{Architectural Backbone.}
To isolate the effects of norm- and residual- design, we instantiate all variants on a shared LLaMA-like backbone.
For sub-1B models, the backbone uses multi-head self-attention (MHA) \citep{attention} with 16 heads, rotary positional embeddings (RoPE) \citep{rope}, RMSNorm \citep{rmsnorm} with $\epsilon=10^{-6}$, SwiGLU feed-forward layers \citep{swiglu}, the GPT-NeoX tokenizer \citep{gpt-neox} with a vocabulary size of 50280, and untied input and output embeddings.
For the 1.6B models, we follow the Qwen3-1.7B attention configuration \citep{qwen3}, replacing MHA with grouped-query attention (GQA) \citep{gqa} using 16 query heads and 8 key-value heads.
All other architectural choices remain unchanged.
Architecture-specific implementation details are provided in Appendix \ref{app:architecture-settings}.

\input{tables/setting_overview}
\input{tables/main_model_configs}

\paragraph{Model Configurations.}
We organize \textsc{DepthBench} into four complementary experimental suites, summarized in Table~\ref{tab:experiment_overview}.
Our main benchmark operates at the 400M total size scale and compares 10 architectures across seven model shapes, spanning aspect ratios from  76.0 ($d_{\mathrm{model}}=1216$, $n_{\mathrm{layer}}=16$) to 9.1 ($d_{\mathrm{model}}=640$, $n_{\mathrm{layer}}=70$).
We additionally conduct an \emph{iso-backbone} control, in which the Transformer backbone size, excluding the input embeddings and LM head, is held fixed at 300M parameters.
This removes a confound of fixed-total-parameter comparisons: narrower models allocate fewer parameters to the embedding and LM head, and thus a larger fraction of the total budget to the Transformer backbone \citep{discrepancies-kaplan-chinchilla}.
To assess robustness across scale, we evaluate three representative aspect ratios at \{200M, 300M, 400M, 500M\} parameters.
Finally, we test whether the same trends persist at the 1.6B scale under three shapes.
400M model configurations are shown in Table \ref{tab:aspect_ratio_configs} and full model configurations for the other three experimental suites are provided in Appendix~\ref{app:sdeep-config}.

For each depth $L=n_{\mathrm{layer}}$, we choose hidden dimension
$d=d_{\mathrm{model}}$
such that the relevant parameter budget is approximately matched.
For sub-1B models, we set the intermediate dimension to
$d_{\mathrm{ff}}
=
\operatorname{ceil}_{16}\left({8d}/{3}\right),
$
where $\operatorname{ceil}_{16}$ denotes rounding up to the next multiple of 16.
For the 1.6B models, we use
$d_{\mathrm{ff}}=3d$.
Since total parameters scale approximately as $N \propto 12Ld^2 + 2Vd$  where $V$ is vocabulary size, increasing depth under a fixed total or backbone parameter budget necessarily requires reducing width. This construction yields a controlled spectrum from shallow--wide to deep--narrow shapes, enabling us to isolate how architectures trade off capacity between width and depth.

\input{tables/optimal_learning_rate}

\subsection{Pre-training Settings} \label{sec:pretraining-settings}

We use OLMo-core\footnote{https://github.com/allenai/OLMo-core} to pre-train all models from scratch on FineWeb-Edu \citep{fineweb}, with a token budget of 20 tokens per parameter following the Chinchilla scaling law \citep{chinchilla} (especially, 8B tokens for 400M models and 32B for 1.6B models), and reserve a disjoint held-out split for evaluation.
We use a sequence length of 2048 and a global batch size of 512 sequences, corresponding to approximately 1M tokens per optimization step.
Optimization uses AdamW \citep{adamw} with $\beta_1=0.9$, $\beta_2=0.95$, $\epsilon=10^{-8}$, weight decay $0.1$, and gradient clipping at $1.0$.
We follow the default OLMo-core initialization with normally distributed weights and standard deviation $0.02$.
The learning rate follows cosine decay with a $10\%$ linear warmup and decays to $10\%$ of its peak value \citep{sgdr}.

For 400M models, we perform architecture-specific learning rate sweeps over
$\{5\times10^{-4}, 1 \times 10^{-3}, 2\times10^{-3}, 5\times10^{-3}\}$,
except for LNS, for which we use
$\{2\times10^{-3}, 5\times10^{-3}, 1 \times 10^{-2}, 2\times10^{-2}\}$.
The optimal learning rate is consistent across width--depth shapes within each architecture and reported in Table \ref{tab:optimal_lr}. We provide the detailed learning rate sweep results  in Appendix \ref{app:lr_sweep}.
We reuse these learning rates for the iso-backbone and 200M--500M multi-scale experiments, and use $5\times10^{-4}$ for 1.6B models following common practice \citep{lns,mix-ln}.
All other pre-training settings are held fixed across architectural variants and aspect ratios.

\section{Main Results} \label{sec:main-results}

\begin{figure}[t]
    \centering
    \includegraphics[width=1.0\linewidth]{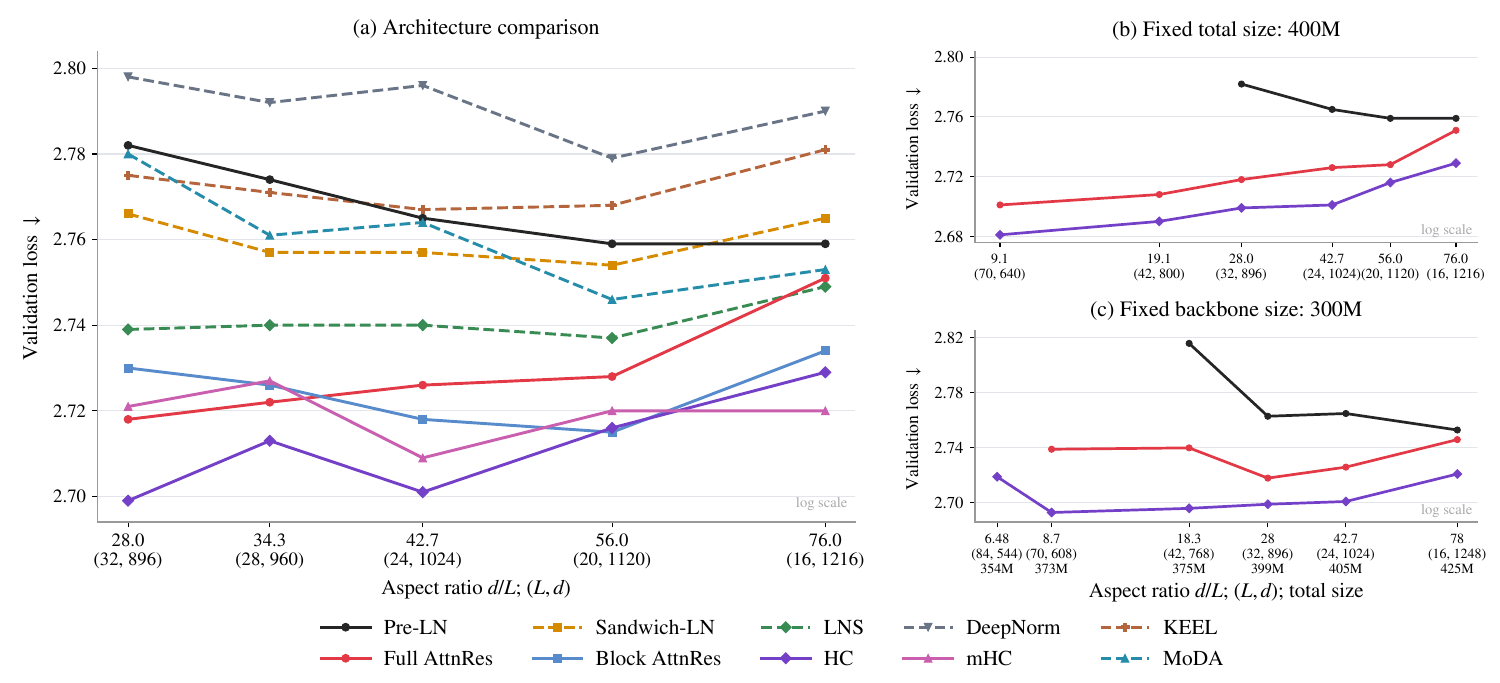}
    \caption{\textbf{Left:} 400M main results of validation loss across 10 architectures. \textbf{Right:} Validation loss across a wider range of shapes, including extreme small aspect ratios, under fixed total size (400M, top) and fixed backbone size (300M, bottom) for Pre-LN, HC and Full AttnRes. }
    \label{fig:main_results_composit}
\end{figure}

\begin{figure}[t]
    \centering
    \includegraphics[width=1.0\linewidth]{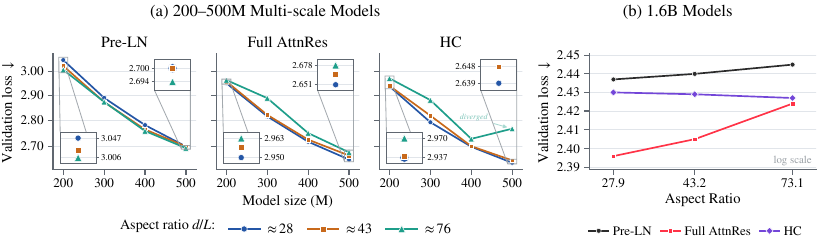}
    \caption{Validation loss for Pre-LN, Full AttnRes, and HC under three aspect ratios. \textbf{Left:}  \{200M, 300M, 400M, 500M\} models. \textbf{Right:} 1.6B models. 
    }
    \label{fig:multiscale}
\end{figure}

\begin{figure}[t]
    \centering
    \includegraphics[width=1.0\linewidth]{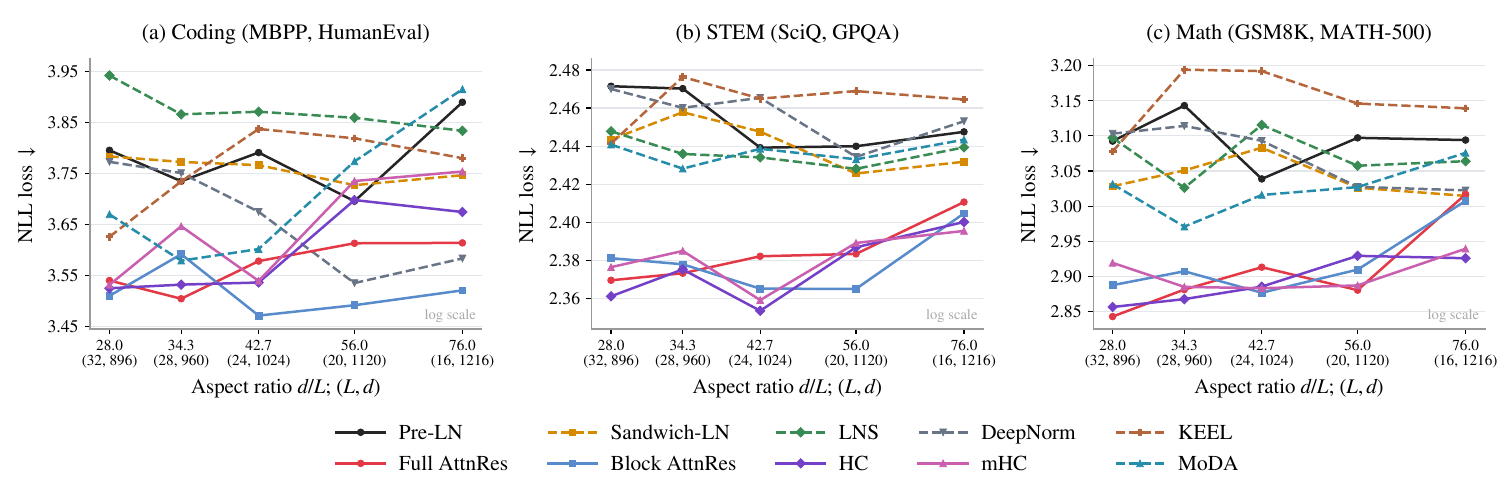}
    \caption{NLL results of 400M models on Coding, STEM and Math categories. We report negative log-likelihood across model aspect ratios and architectures. Each panel shows the average NLL over two representative tasks: MBPP \citep{mbpp} and HumanEval \citep{humaneval} for \textbf{coding (left)}, SciQ \citep{sciq} and GPQA \citep{gpqa} for \textbf{STEM (middle)}, and GSM8K \citep{gsm8k} and MATH-500 \citep{math500} for \textbf{math (right)}.
    }
    \label{fig:nll_aspect_ratio_scaling}
\end{figure}

Figure~\ref{fig:main_results_composit} shows validation loss on  the 400M benchmark across a wide range of
architectures and aspect ratios. Figure \ref{fig:multiscale} shows validation loss across aspect ratios of Pre-LN, Full AttnRes and HC in 200M--500M and 1.6B suites. To complement the  validation loss, we further evaluate teacher-forced negative log-likelihood (NLL) on coding, STEM, and math
tasks in Figure~\ref{fig:nll_aspect_ratio_scaling}, following the NLL-based
pre-training evaluation protocol  in
\citep{mai, nexus}. We observe a clear pattern:
{whether increasing depth helps strongly depends on the residual connections.}

\paragraph{Pre-LN and normalization variants are largely insensitive or even unfavorable to increasingly deep–narrow shape.} As shown in Figure \ref{fig:main_results_composit} (a), for Pre-LN, validation loss monotonically increases from 2.759 at $L=16$ to 2.782 at
$L=32$, showing that reallocating parameters from width to depth hurts
performance. Sandwich-LN, LNS, DeepNorm, KEEL and MoDA also show either weak or
non-monotonic trends, with their optima occurring at
intermediate or shallower shapes.

\paragraph{HC and Full AttnRes scale favorably with depth and even surprisingly continue to improve at extremely deep shapes.}
In contrast, both HC and Full AttnRes consistently improve as models become deeper
and narrower. Full AttnRes steadily  improves from 2.751 at $L=16$ to 2.718 at $L=32$, while HC
improves from 2.729 at $L=16$ to 2.699  at $L=32$.
This trend is the opposite of Pre-LN, suggesting that these new residual designs
strongly prefer deep shapes.

We further extend the scaling range, up to 70 layers. As shown in Figure \ref{fig:main_results_composit} (b-c), the favorable trend persists in both fixing total size and backbone size. Notably, with the backbone size fixed, deep--narrow models improve even as the total size decreases. This suggests that their depth-scaling gains arise from the width--depth allocation itself rather than simply from increased backbone size. Overall, HC and Full AttnRes are exceptionally well suited to depth scaling, with performance continuing to improve as models are pushed toward increasingly deep--narrow, even extreme, shapes.

\paragraph{The favorable width-depth scaling behavior of HC and Full AttnRes is less evident in their derived variants, Block AttnRes and mHC.} As shown in Figure \ref{fig:main_results_composit} (a), Block AttnRes remains competitive but performs best at an intermediate shape, reaching 2.715 at $L=20$ before degrading to 2.730 at $L=32$.
Similarly, mHC achieves strong overall performance and its best performance at $L=24$ but shows no consistent
gain with increasing depth. 

\paragraph{A wide range of scales coupled with domain-specific evaluation shows the consistent trend.} As shown in Figure \ref{fig:multiscale}, across 200M–500M models, Full AttnRes consistently benefits from deeper–narrower shapes, and this trend persists at 1.6B, suggesting that its favorable depth scaling extends beyond the 400M regime. HC shows a similar trend at smaller scales, but exhibits less stable behavior as scale increases: the 500M run at the largest aspect ratio encounters gradient explosion, while the 1.6B results do not show the same clear improvement with depth. The reason is  that HC is more sensitive to optimization hyperparameters and may require finer learning-rate tuning across scales, consistent with the motivation of mHC to improve the large-scale optimization stability of unconstrained HC \citep{hc, mhc}.

As shown in Figure~\ref{fig:nll_aspect_ratio_scaling}, deeper HC and Full AttnRes models
generally achieve lower NLL across coding, STEM, and math evaluations.
The trend is particularly clear for STEM and math, where both architectures
consistently improve as capacity is shifted toward depth.
In contrast, Pre-LN and most other variants show weaker or non-monotonic
 trends.
These results suggest that the favorable width-depth scaling
translates to domain-specific predictive capability rather than only lower
held-out pre-training loss.

Overall, our results identify HC and Full AttnRes as the two architectures with
the clearest favorable depth-scaling behavior.
We next study how these architectures utilize their layers.

\section{Analysis} 

\subsection{Evaluating Depth Utilization Across Architectures} \label{sec:depth-metric}
Prior work has shown that deep layers in Pre-LN Transformers can become
increasingly ineffective \cite{lns,alphaq}.
In particular, representations produced by neighboring deep layers become
increasingly similar \cite{lns, inversedepthscaling}, suggesting that many layers only make small
refinements to the residual stream \cite{depth-effective}.
We therefore ask whether the favorable depth scaling is
accompanied by more effective utilization across their layers.

\begin{figure}[t]
    \centering
    \includegraphics[width=1.0\linewidth]{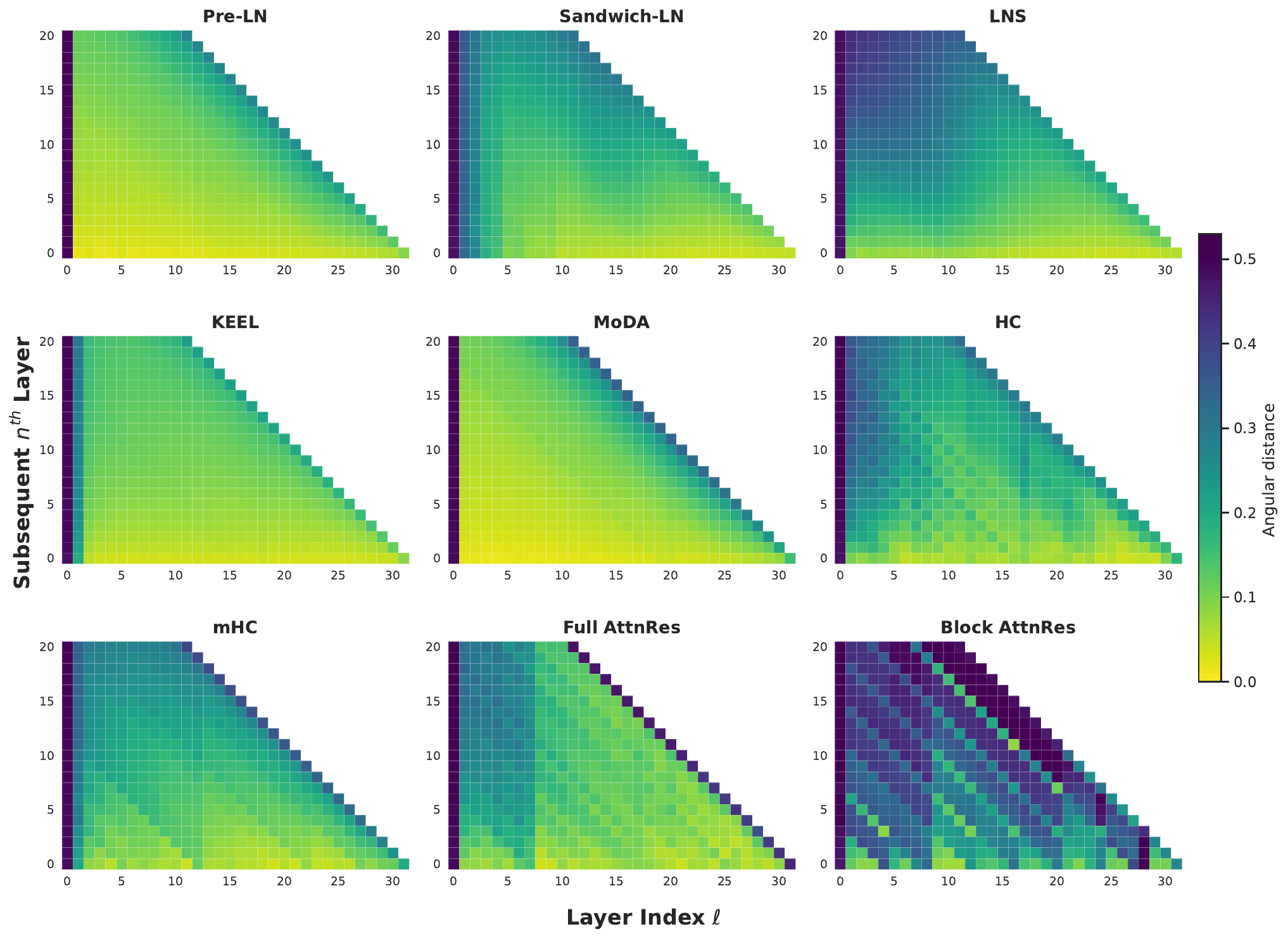}
    \caption{\textbf{Angular distance from the initial layer $\ell$ (x-axis) and its subsequent $n^{th}$ layer (y-axis)}. Angular distance ranges in [0, 1] and yellow indicates smaller distances and dark blue indicates larger distances. Here, we use a unified learning rate of $2\times10^{-3}$ for all models (only LNS does not use the optimal learning rate), because the learning rate significantly affects the learned representations, as shown in Appendix \ref{app:lr_sweep}. We exclude DeepNorm because it explodes at a learning rate of $2\times10^{-3}$.}
    \label{fig:l32_angular_distance}
\end{figure}

\paragraph{Representation diversity across depth.}
We first examine the \textit{angular distance} between representations at
different depths, following prior work \cite{lns, mix-ln}.
For representations $h_\ell$ and $h_{\ell+n}$ after the outputs\footnote{The {representation} here corresponds to $h_{\ell}$ or $H_{\ell}$ in Table~\ref{tab:architectures}. For HC and mHC, the outputs from the four streams are concatenated together; for AttnRes, we use the output after each depth mixing.} from layers $\ell$ and
$\ell+n$, respectively, the angular distance is defined as:
\begin{equation}
d(h_\ell, h_{\ell+n})
=
\frac{1}{\pi}
\arccos
\left(
\frac{
h_\ell \cdot h_{\ell+n}
}{
\|h_\ell\|_2 \|h_{\ell+n}\|_2
}
\right).
\end{equation}
A smaller angular distance indicates more similar representations and therefore
less representational change.

As shown in Figure~\ref{fig:l32_angular_distance}, The hidden representations of Pre-LN, Sandwich-LN, LNS, DeepNorm and KEEL become broadly similar across depth, generally similar with patterns in \cite{lns, mix-ln}. In contrast, HC- and AttnRes- architectures exhibit not only larger angular distances, but
more importantly, a markedly {non-smooth} structure across depth.
Their representations continue to change in a layer-specific and non-uniform
manner, rather than gradually converging toward small refinements of the same
underlying representation.
Block AttnRes exhibits a distinct {block-structured} pattern:
representations remain relatively smooth within each block, but change sharply
across block boundaries.
mHC shows a more moderate non-smooth pattern, but remains clearly distinct from
the progressively smoothed representations like Pre-LN.

The key distinction is therefore not merely the magnitude of representational
change, but its structure across depth: HC, mHC and AttnRes preserve
heterogeneous, layer-specific transformations, whereas other architectures progressively
collapses toward smoother and increasingly similar deep-layer representations.

\paragraph{Layer perturbations across depth.}
We further study
whether each layer is functionally important by adopting \textit{causal scores} and \textit{permutation scores} from \cite{sparsity4cod, depth-effective} to quantify how strongly individual layers
affect subsequent computation and how interchangeable different layers are. For a skipped layer $s$ and a subsequent layer $\ell > s$, the causal score is
defined as:
\begin{equation}
C(s,\ell)
=
\frac{
\left\|
(h_{\ell+1}-h_{\ell})
-
(\bar{h}_{\ell+1}-\bar{h}_{\ell})
\right\|_2
}{
\|h_{\ell+1}-h_{\ell}\|_2
},
\end{equation}
where $h$ denotes the original hidden states and $\bar{h}$ denotes the hidden
states obtained after skipping layer $s$.
A larger $C(s,\ell)$ means that removing layer $s$ more strongly changes the
update performed by a subsequent layer $\ell$, indicating stronger causal
dependence across depth.

We additionally measure how sensitive the model is to exchanging the order of
two layers.
For layers $\ell_1$ and $\ell_2$, the permutation score is defined as:
\begin{equation}
P(\ell_1,\ell_2)
=
\frac{
\left|
\mathcal{L}(M)
-
\mathcal{L}(M_{\mathrm{swap}(\ell_1,\ell_2)})
\right|
}{
\mathcal{L}(M)
},
\end{equation}
where $\mathcal{L}(M)$ denotes the language modeling loss of the original
model and $\mathcal{L}(M_{\mathrm{swap}(\ell_1,\ell_2)})$ denotes the loss of the model after swapping
layers $\ell_1$ and $\ell_2$.
A larger permutation score indicates that exchanging the two layers causes a
larger performance degradation, suggesting that they perform more specialized
and order-dependent computations.

\begin{figure}
    \centering
    \includegraphics[width=1.0\linewidth]{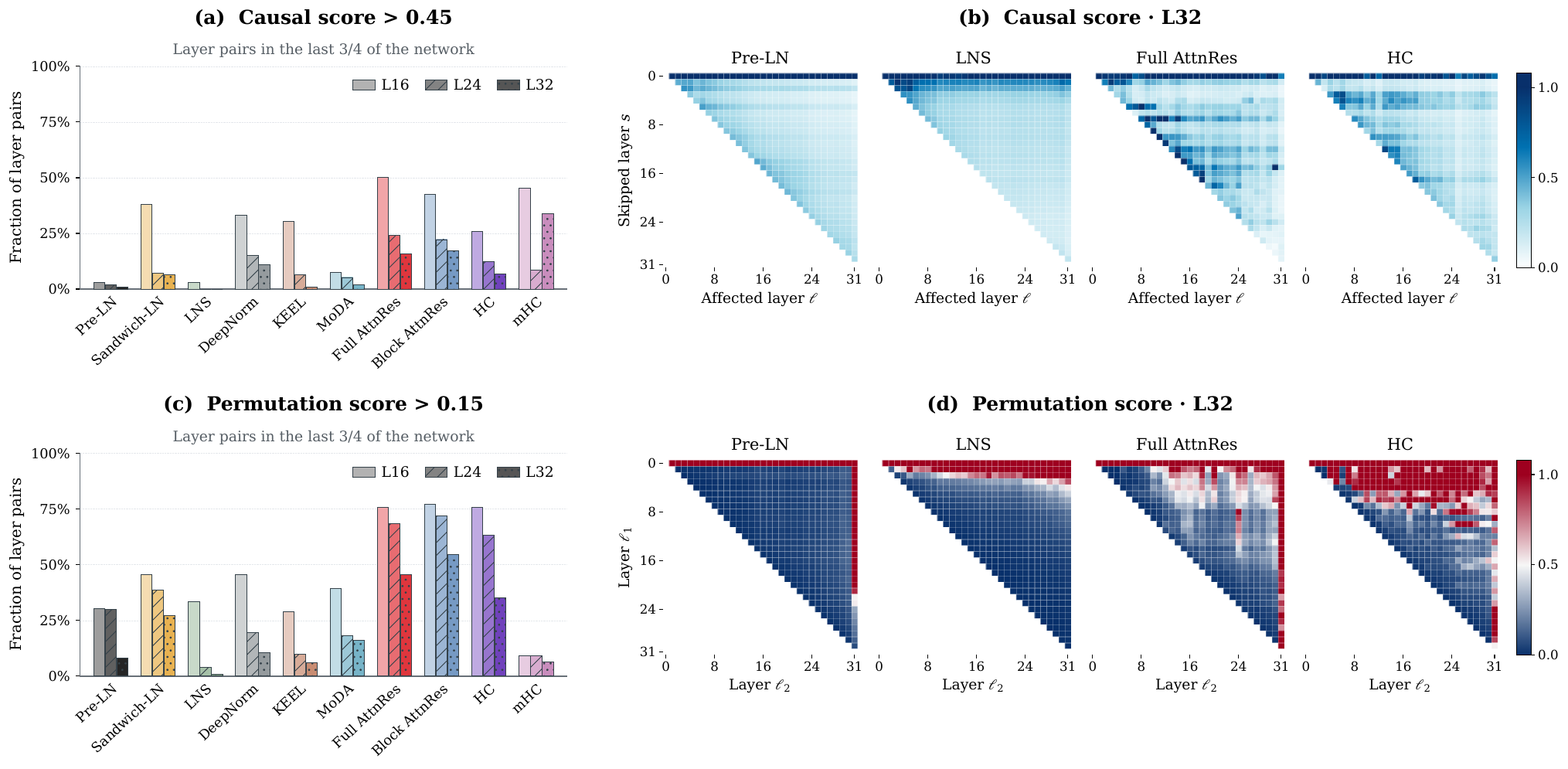}
    \caption{\textbf{(a,c)} Fraction of layer pairs in the last $3/4$ of the network exceeding the causal-score ($>0.45$) and permutation-score ($>0.15$) thresholds, respectively.
    \textbf{(b,d)} Pairwise causal and permutation score maps for $L=32$ models.
    Higher causal scores indicate stronger cross-layer dependence, while higher permutation scores indicate lower layer interchangeability.}
    \label{fig:causal_permutation_main}
\end{figure}

While we use the pairwise scores above, our aggregation differs from that
of~\cite{sparsity4cod}.
We find that the first few layers are typically important across almost all
architectures, which can dominate a global layer effectiveness statistic and
obscure differences in how later layers are utilized.
Since our focus is specifically on effective \emph{deep} computation, we
therefore restrict the analysis to the last three quarters of the network and
measure the fraction of pairwise scores that exceed a fixed threshold. Specifically, let
\begin{equation}
\mathcal{S}_{\mathrm{late}}
=
\left\{
(s,\ell):
s \geq \left\lceil L/4 \right\rceil,\;
\ell > s
\right\}
\end{equation}
denote the causal pairs whose skipped layer lies in the last three quarters of
the network.
We summarize causal utilization as:
\begin{equation}
R_{\mathrm{causal}}
=
\frac{1}{|\mathcal{S}_{\mathrm{late}}|}
\sum_{(s,\ell)\in\mathcal{S}_{\mathrm{late}}}
\mathbf{1}\left[C(s,\ell) > \tau_c\right],
\qquad
\tau_c = 0.45.
\end{equation}
Similarly, for layer pairs
\begin{equation}
\mathcal{P}_{\mathrm{late}}
=
\left\{
(\ell_1,\ell_2):
\ell_1 \geq \left\lceil L/4 \right\rceil,\;
\ell_2 > \ell_1
\right\},
\end{equation}
we define:
\begin{equation}
R_{\mathrm{perm}}
=
\frac{1}{|\mathcal{P}_{\mathrm{late}}|}
\sum_{(\ell_1,\ell_2)\in\mathcal{P}_{\mathrm{late}}}
\mathbf{1}\left[P(\ell_1,\ell_2) > \tau_p\right],
\qquad
\tau_p = 0.15.
\end{equation}
These statistics directly measure how frequently later layers exhibit
substantial causal influence or non-interchangeable computation, while avoiding
the universally strong effects of the earliest layers.

As shown in Figure~\ref{fig:causal_permutation_main}, Pre-LN and its normalization variants, as well as MoDA, exhibit very few
strong causal interactions in its deeper layers.
For the $L=32$ model, only about $1\%$ of Pre-LN's late-layer causal scores exceed
$0.45$, compared with roughly $16\%$ for Full AttnRes and $7\%$ for HC.
The separation is even clearer under layer permutation: only about $8\%$ of
Pre-LN permutation scores exceed $0.15$, compared with approximately $46\%$
for Full AttnRes and $35\%$ for HC. Although focusing on the last three quarters could potentially benefit
Post-LN-based methods such as DeepNorm and KEEL, these methods still do not
exhibit larger causal or permutation scores in our analysis. mHC likewise shows a marked drop in
permutation sensitivity, consistent with its weaker width-depth scaling.
The corresponding pairwise heatmaps show the same pattern, with substantially
stronger cross-layer dependence and order sensitivity in Full AttnRes and HC. Complete visualizations are provided in Appendix \ref{heatmaps}.

Taken together, 
Pre-LN,  most normalization-based variants and MoDA become broadly smooth
across depth, with weak causal influence and limited sensitivity to
layer permutation, indicating that many late layers contribute only small and
partly substitutable refinements.
In contrast, HC and Full AttnRes maintain non-smooth, layer-specific
representational changes together with substantially stronger causal dependence
and permutation sensitivity, suggesting that their additional layers remain
functionally consequential.

\subsection{Characterizing Depth Processing in Full AttnRes and HC} \label{sec:depth-processing}

Section~\ref{sec:depth-metric} shows that  AttnRes and HC retain stronger
cross-layer dependence and order sensitivity than Pre-LN and its normalization variants. We next examine whether
these architectures realize more effective depth through different computational mechanism.

\begin{figure}[t]
    \centering
    \includegraphics[width=1.0\linewidth]{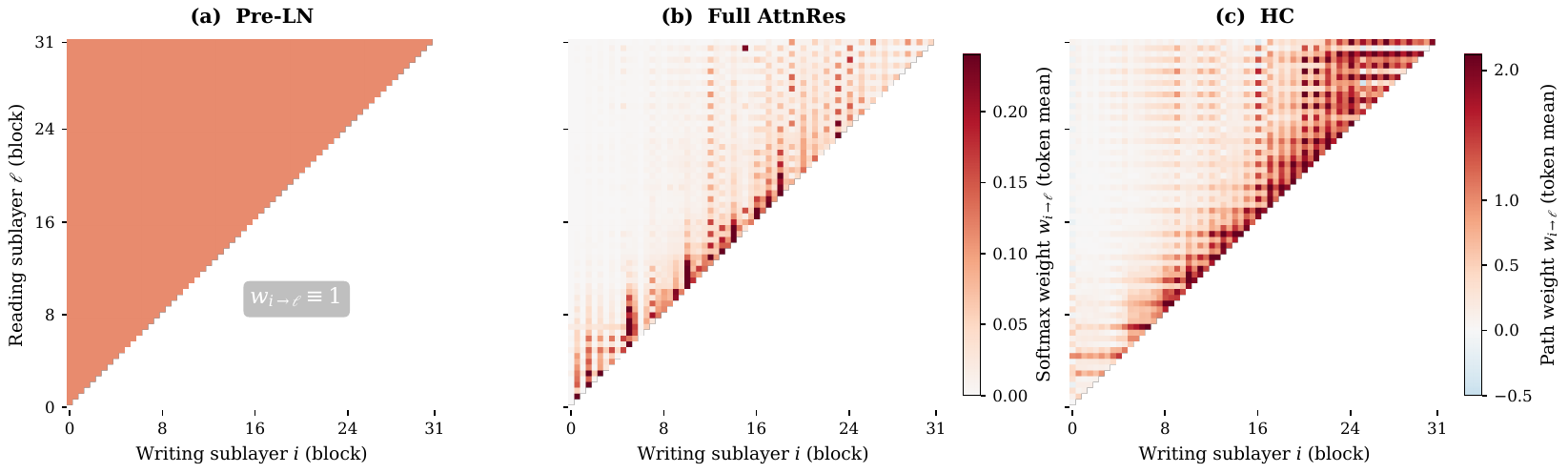}
    \caption{\textbf{Explicit residual-path weights $w_{i\to\ell}$.}
Each cell shows the mixing weight of sublayer output $f_i$ to the pre-norm input of a later sublayer $\ell$, averaged over validation calibration tokens.
\textbf{(a) Pre-LN:} $w_{i\to\ell}\equiv1$.
\textbf{(b) Full AttnRes:} $w_{i\to\ell}$ is the per-token softmax weight on $f_i$. 
\textbf{(c) HC:} $w_{i\to\ell}=\beta_i^\top A_{i+1}\cdots A_{\ell-1}\alpha_\ell$, combining write mappings, residual mappings, and read mappings.}
    \label{fig:pathway_weights}
\end{figure}

\paragraph{Explicit paths across depth.}
Figure~\ref{fig:pathway_weights} traces how earlier sublayer outputs contribute to
later computations. Pre-LN accumulates all updates in a single residual stream,
giving $w_{i\rightarrow\ell}\equiv1$. Full AttnRes instead forms a normalized,
non-negative mixture over the embedding and all preceding sublayer outputs,
providing direct access to earlier computations. HC realizes cross-depth access
recursively, with the contribution of branch output $f_i$ to branch $\ell$ given by
$
w_{i\rightarrow\ell}
=
\beta_i^{\top}A_{i+1}\cdots A_{\ell-1}\alpha_\ell .
$
Its learned paths are strongest locally but remain substantial over longer ranges.
Thus, Full AttnRes directly retrieves stored earlier outputs, whereas HC propagates
and recombines them through interacting residual streams.

\begin{figure}[t]
    \centering
    \includegraphics[width=1.0\linewidth]{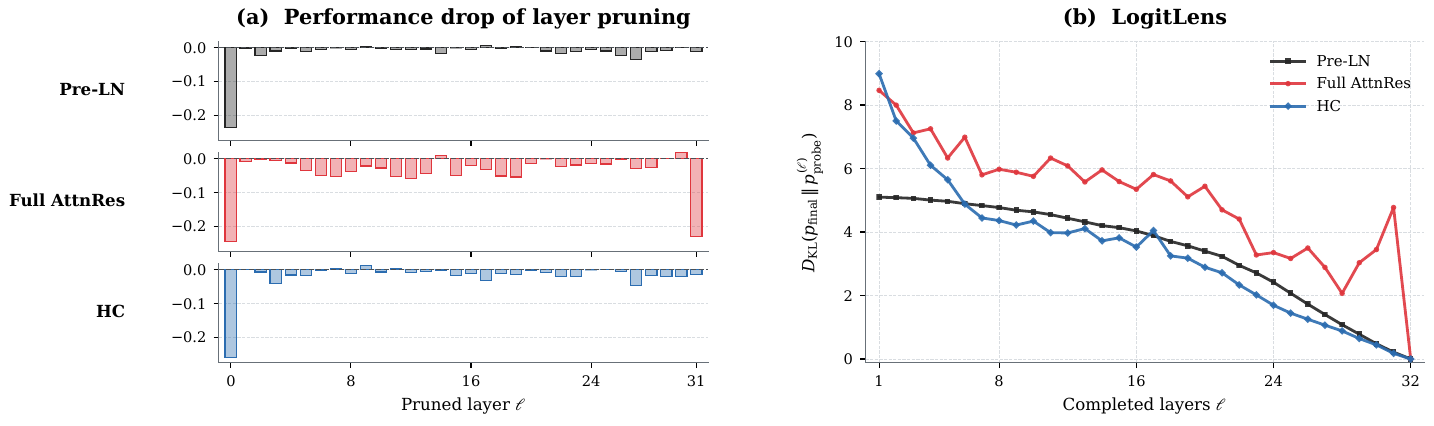}
    \caption{\textbf{Left:} Performance drop after removing a single layer.
    \textbf{Right:} The KL divergence between layerwise and final prediction distributions. }
    \label{fig:layer_pruning}
\end{figure}

\paragraph{Functional consequences.}
We first prune individual layers and measure the resulting change in ARC-Easy
accuracy (Figure~\ref{fig:layer_pruning} (a)). Pre-LN and HC show similar patterns:
both are dominated by the first layer, while pruning most later layers causes only
modest degradation. Full AttnRes is sensitive across a broader range, with strong
dependence on both its first and final layers.

We further apply LogitLens \citep{depth-effective} to track how predictions evolve
across depth. For Pre-LN, we decode the current residual stream; for Full AttnRes,
the mixture produced by the next layer's depth-mixing query; and for
HC\footnote{We implement HC following the original HC paper \citep{hc}, where the final output is an
sum of the four residual streams. Later HC/mHC implementations, such as
DeepSeek V4 \citep{v4}, may instead use a learnable final mixer.}, the sum of
its four residual streams. For probe state $h_\ell$, we compute
$
p_{\mathrm{probe}}^{(\ell)}
=
\operatorname{softmax}
\!\left(
W_{\mathrm{LM}}\operatorname{Norm}(h_\ell)
\right)
$
and report
$D_{\mathrm{KL}}(p_{\mathrm{final}}\Vert p_{\mathrm{probe}}^{(\ell)})$.
As shown in Figure~\ref{fig:layer_pruning} (b), Pre-LN approaches the final
prediction monotonically, consistent with progressive refinement of a single
residual stream. HC follows a similarly smooth trajectory, but its earlier
representations remain farther from the final prediction. Full AttnRes is markedly
less monotonic, consistent with continued retrieval and late integration of stored
sub-layer output sources.

Together, these results suggest that Full AttnRes and HC need not make every layer
individually indispensable. Rather, their learned cross-depth routing preserves
differentiated access to earlier computations, maintaining stronger cross-layer
dependence and order sensitivity instead of reducing later layers to largely
interchangeable refinements.

\subsection{Potential Hidden Cost for mHC and Block AttnRes} \label{app:hidden_cost}
Figure \ref{fig:main_results_composit} shows that the  scaling behavior of Full AttnRes and HC is less evident in their derived variants, Block AttnRes and mHC.  We investigate the underlying causes of this discrepancy.

\begin{figure}
    \centering
    \includegraphics[width=1.0\linewidth]{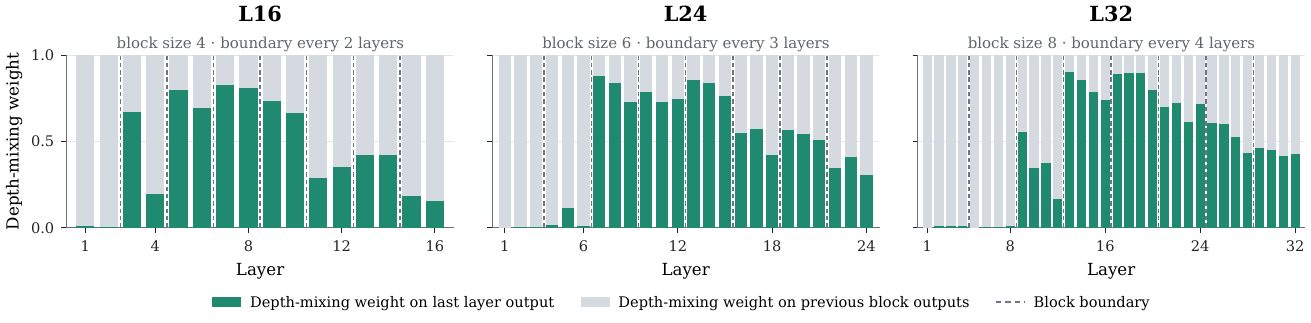}
    \caption{\textbf{Block AttnRes depth-mixing mechanism
  at $L=16$, $24$, $32$.}
  At each layer, the share of the depth-mixing weight in
  Eq. ~\ref{eq:block_attnres_mix} placed on the last layer output
  ($\alpha_\ell$, green) versus on previous block outputs
  ($\sum_{\mathbf{v}}\beta_{\ell,\mathbf{v}}$, grey); dashed lines mark block
  boundaries, where the set of saved block outputs is updated.}
    \label{fig:block_attnres}
\end{figure}

\paragraph{Block AttnRes.}
Block AttnRes restricts the sources for depth-mixing to a slower growing schedule.
Layer outputs are accumulated into a running sum $\mathbf{s}_\ell$, and only at
a \emph{block boundary}, every $b/2$ layers for block size $b$, is the running
sum appended to the set $\mathcal{B}_\ell$ of saved block outputs that later
layers may attend over and then reset.
At every layer the depth-mixing softmax is taken over
$\{\mathbf{s}_\ell\}\cup\mathcal{B}_\ell$, so the mixed input is
\begin{equation}
  \tilde{\mathbf{h}}_\ell
  \;\propto\;
  \alpha_\ell\,\mathbf{s}_\ell
  \;+\;
  \sum_{\mathbf{v}\in\mathcal{B}_\ell}\beta_{\ell,\mathbf{v}}\,\mathbf{v},
  \qquad
  \alpha_\ell + \textstyle\sum_{\mathbf{v}}\beta_{\ell,\mathbf{v}} = 1,
  \label{eq:block_attnres_mix}
\end{equation}
where $\alpha_\ell$ is the weight on the \emph{last layer output} (the running
sum) and the $\beta_{\ell,\mathbf{v}}$ are the weights on \emph{previous block
outputs}.
We follow original AttnRes \citep{attnres} and Kimi K3 \citep{k3} settings to fix the number of boundaries at 8, and the block
size grows with depth ($b=4,5,6,7, 8$ for $L=16,20, 24,28, 32$).
Figure~\ref{fig:block_attnres} shows how this schedule shapes the
computation, by plotting $\alpha_\ell$ for the pre-trained models.

Before the first block output has been saved, the depth-mixing softmax assigns
essentially zero weight to the running sum: each layer's own computation is dropped from
the mixed input, and the network effectively keeps re-reading the embedding.
Only once saved block outputs exist does $\alpha_\ell$ rise.
The length of this initial dead segment scales with depth ($2$, $5$, and $8$
layers at $L=16$, $24$, and $32$), so a quarter of the $L=32$ network
can not benefit from stacking layers. This potentially explains why Block AttnRes stops benefiting from added depth.


\begin{figure}[t]
    \centering
    \includegraphics[width=\linewidth]{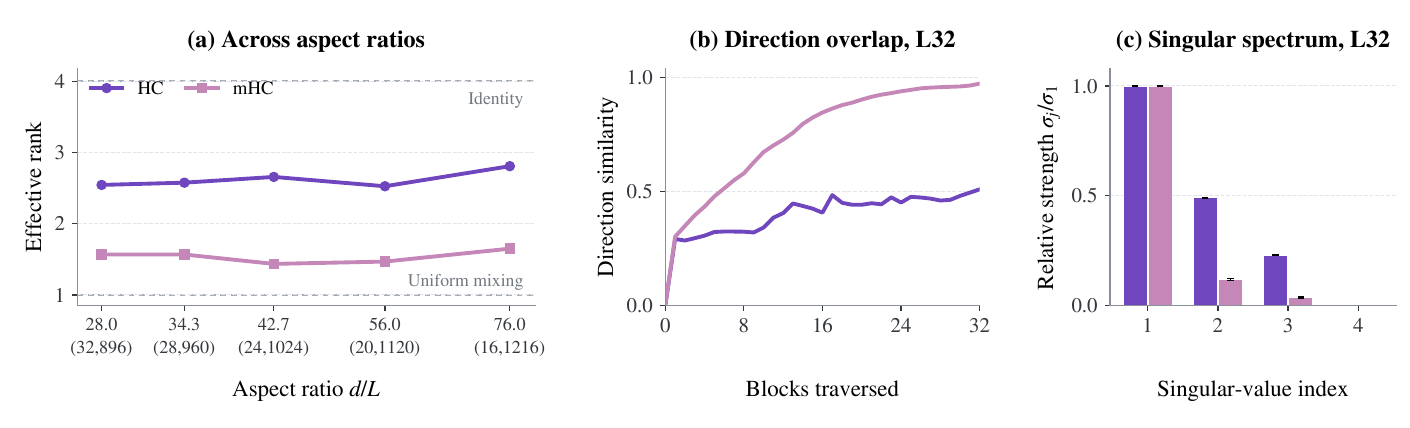}
    \caption{\textbf{Residual transport in HC and mHC.}
    \textbf{(a)} Effective rank across aspect ratios, with shapes $(L,d)$ below
    each tick. Effective rank is the exponential of the entropy of the
    normalized singular values.
    \textbf{(b)} Mean absolute cosine similarity between propagated stream
    directions in $L=32$ models.
    \textbf{(c)} Singular values of their full residual products, normalized
    by the largest.
    Statistics are averaged over sampled held-out FineWeb-Edu data;
    bands and error bars show 95\% document-bootstrap intervals.}
    \label{fig:hc_mhc_residual_rank}
\end{figure}

\paragraph{mHC.}
mHC retains HC's multi-stream structure but constrains its read/write
weights to be non-negative and uses Sinkhorn--Knopp normalization to make
residual mappings approximately doubly stochastic~\citep{mhc}.
These constraints aim to stabilize propagation across depth.
However, as shown in Figure \ref{fig:main_results_composit},
mHC performs best at $L=24$ and does not retain HC's gains in deeper,
narrower models.

To examine this difference, we compose all attention and FFN residual
maps along each forward pass.
As shown in Figure~\ref{fig:hc_mhc_residual_rank}(a), the resulting products
have consistently lower effective rank in mHC: $1.44$--$1.65$, compared
with $2.53$--$2.81$ for HC.
At $L=32$, mHC's propagated stream directions become nearly collinear,
with much weaker non-leading singular values, as shown in
Figure~\ref{fig:hc_mhc_residual_rank}(b--c).

The doubly stochastic constraint suggests a mechanism for this
concentration. Exact doubly stochastic maps share the uniform mixing
matrix as a fixed point:
\begin{equation}
    U=\frac{1}{m}\mathbf{1}\mathbf{1}^{\top},
    \qquad A_\ell U=UA_\ell=U.
    \label{eq:mhc_uniform_mixing}
\end{equation}
When repeated mixing contracts the remaining stream directions,
residual products approach $U$, making earlier contributions less
distinguishable across streams even as new branch outputs enter.
Different sufficiently mixing orders of fixed residual maps then
approach the same average, consistent with mHC's weaker permutation
effects as shown in Figure~\ref{fig:causal_permutation_main}.
Together, these observations suggest a trade-off: stabilizing residual
transport can reduce the diversity of long-range routes through which
additional layers reuse earlier computations.

\begin{figure}[t]
    \centering
    \includegraphics[width=1.0\linewidth]{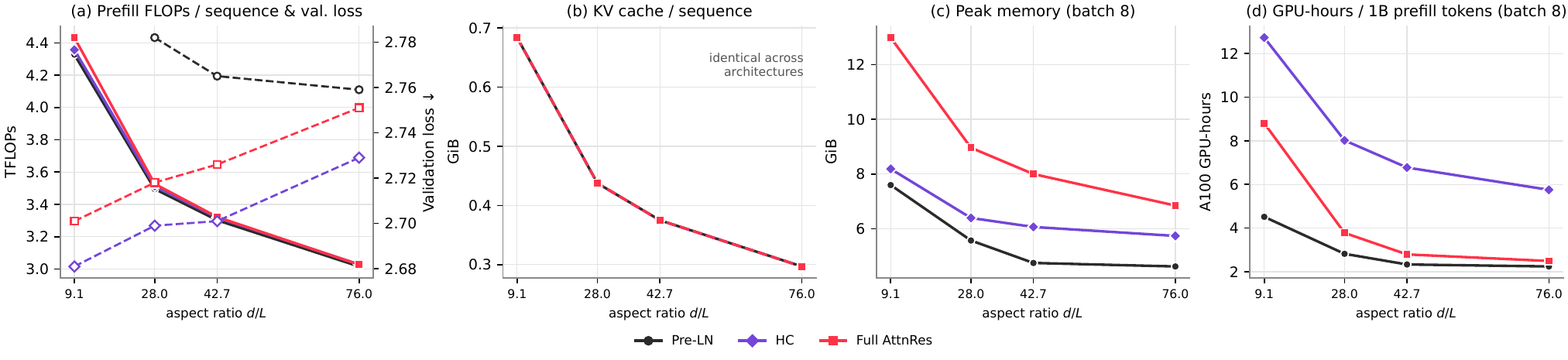}
    \caption{\textbf{Measured prefill cost} including FLOPs, KV cache memory, Peak GPU memory and GPU hours at sequence length $=4096$, model size at about 400M, BF16, A100-SXM4-40GB. The dashed line in (a) corresponds to validation loss.
  }
    \label{fig:prefill_cost}
\end{figure}

\subsection{No Free Lunch: The Accuracy--Efficiency Trade-off of Depth Scaling} \label{sec:efficiency-trade-off}

Depth scaling is not free even under a fixed parameter budget. Since
$N \approx 12Ld^2 + 2Vd$, while the dominant projection cost proportional to $Ld^2$ remains approximately
constant, quantities proportional to $Ld$ increase when scaling depth. For example, during prefill
with sequence length $T$, attention prefill cost scales as $\mathcal{O}(T^2Ld)$ and
becomes increasingly important in the long-context regime $T \gg d$, while KV cache
memory similarly scales as $\mathcal{O}(TLd)$. We provide the FLOPs counting analysis in Appendix~\ref{app:flops}.

Moreover, the practical latency overhead can exceed the increase suggested by FLOPs alone, as deeper models incur lower hardware utilization.  Figure~\ref{fig:prefill_cost} reports
prefill cost at sequence length $T=4096$ for our 400M models in BF16 on
A100-SXM4-40GB GPUs under specific setting. We measure forward FLOPs and KV-cache size per sequence,
as well as peak memory and GPU-hours for a batch size of 8. 
A first observation is that depth itself introduces a non-negligible
systems cost even at a fixed parameter scale. As the aspect ratio decreases
from 76.0 to 9.1, prefill FLOPs increase from roughly 3.0 to 4.4 TFLOPs per
sequence, while the KV cache increases by more than $2\times$. The latter is
identical across architectures, confirming that this component is intrinsic
to the width--depth reallocation rather than to a particular residual design.
More importantly, the practical cost of depth grows substantially
fast. Peak memory and GPU-hours increase sharply
toward the deepest configurations. Full AttnRes has the largest peak-memory
footprint, consistent with its explicit cross-layer source access, whereas HC
shows the highest measured GPU-hour cost despite having a much smaller
difference in nominal FLOPs. 

Notably, those algorithmic and empirical forward costs do not directly reflect  training wall clock. Deeper models incur more sequential execution and smaller matrix multiplications, so their practical cost depends strongly on hardware, kernels, parallelism, and memory behavior. When fixing parameter counts, depth scaling therefore introduces a systems-level efficiency trade-off rather than a universally fixed compute penalty. HC and Full AttnRes obtain better
modeling performance from additional depth, but realizing these gains at scale
likely requires architecture--system co-design. In particular, specialized kernels and infrastructure-level aspecs such as pipeline scheduling, communication overlap among others might shape the practical frontier. To facilitate further exploration of these systems-level trade-offs, we release our codebase together with checkpoints spanning the wide range of model aspect ratios.

\section{Related Work}

\paragraph{Scaling Laws and Model Aspect Ratios.}
Scaling laws have primarily characterized language modeling performance as a
function of model size, data, and training compute \cite{kaplan, chinchilla}. 
Within the regimes considered by early studies, architectural choices such as
network width and depth were found to have comparatively limited effects once
model size was controlled~\cite{kaplan}.
However, subsequent work has shown that how capacity is allocated across
architectural dimensions can matter. \citet{depthlimits} 
characterizes distinct depth-efficient and depth-inefficient regimes in
self-attention. \citet{depth-pretrain-downstream} and \citet{depth-impact-generalization} show that model shape can substantially affect downstream fine-tuning performance, implying that deep models potentially have better generalization.
Relatedly, MobileLLM~\cite{mobilellm} demonstrates the practical
effectiveness of deep-and-thin architectures in the sub-billion-parameter
regime. More recently, Gemstones~\cite{gemstones} systematically studies
scaling across diverse model shapes and shows that scaling-law prescriptions
can be sensitive to width--depth choices and experimental design.

These studies, however, largely examine model shape on fixed conventional model architectures.
In contrast, we study width--depth allocation across different normalization and residual
architectures and show that whether reallocating capacity from width to depth
is beneficial depends strongly on residual connection design.

\paragraph{Deep Transformer Training and the Curse of Depth.}
The original Post-LN Transformer \cite{post-ln} can exhibit unstable gradients at
initialization, while Pre-LN substantially improves optimization stability by
providing better-behaved gradient propagation \cite{pre-ln}. Subsequent work further improves the trainability of deep Transformers through normalization, residual scaling, and initialization. For example, DeepNorm \cite{deepnorm} combines residual scaling with
depth-aware initialization to stably train Post-LN Transformers with up to thousands
of layers, while Sandwich-LN \cite{sandwich-ln1, sandwich-ln2} improves activation-variance
control and gradient stability through putting
normalization both before and after the transformation branch.

More recent work has shifted attention from merely \emph{training} deep models
to whether their additional layers are effectively \emph{utilized}. Mix-LN~\cite{mix-ln} shows that Pre-LN can produce increasingly weak
gradients in deeper layers and combines Pre- and Post-LN to improve their
contribution.
Similarly, the Curse of Depth~\cite{lns} attributes ineffective deep
layers in Pre-LN Transformers to depth-dependent residual-stream variance growth and
introduces LayerNorm Scaling to strengthen deep-layer updates. Recent approaches further revisit the Pre-/Post-Norm trade-off:
KEEL~\cite{keel} stabilizes very deep Post-LN Transformers through
highway-style residual connections, SiameseNorm~\cite{siamesenorm}
decouples Pre- and Post-Norm behavior using two coupled streams, and
KiteNorm~\cite{kitenorm} regularizes hidden-state variance to
stabilize Post-LN training.

These studies establish that trainability and depth utilization are distinct:
a Transformer can be optimized successfully at large architectural depth while
still deriving limited benefit from its later layers.
Our work takes the next step by asking whether improved deep-layer utilization
is sufficient to make depth a favorable scaling axis under a fixed parameter
budget.

\paragraph{Residual Connections.}
Residual connections are a fundamental mechanism for enabling information and
gradient propagation through deep networks.
Highway Networks \cite{highwaynet} introduced gated shortcuts to
regulate information flow across layers, while ResNet \cite{resnet}
established identity residual connections as a simple and effective means of
optimizing substantially deeper networks.
Beyond local identity shortcuts, DenseNet \cite{densenet} showed the
benefit of providing layers with direct access to earlier representations,
motivating richer forms of cross-depth information reuse.

Recent work has generalized this idea in Transformers by making residual
propagation increasingly learnable.
DenseFormer \cite{denseformer} mixes current and preceding
representations through learned depth-weighted averaging, while
LAuReL \cite{laurel} augments the canonical residual layer with
learnable residual mappings.
Hyper-Connections (HC) \cite{hc}  expand the
single residual stream into multiple streams with learned read, write, and
mixing operations. mHC \cite{mhc} further constrains these mappings to
restore the identity-preserving properties required for stable large-scale
training.
Complementarily, AttnRes \cite{attnres} replace
fixed residual accumulation with input-dependent attention over preceding
layer outputs, enabling selective access to computations across depth, and
Block AttnRes trades this fine-grained access for a more efficient block-wise
variant.
Related approaches such as MoDA \cite{moda} and Depth-Attention \cite{depth-attn} provide direct cross-layer
access through depth-wise key--value representations.

While these methods demonstrate that richer residual pathways can improve
optimization and model quality, they are typically evaluated under different
model shapes and training settings.
We instead study them under controlled width--depth aspect ratio scaling, asking whether
their residual topology actually converts additional architectural depth into
effective computation.

\section{Conclusion}
In this work, we introduce \textsc{DepthBench}, a controlled benchmark for studying when architectural depth translates into effective computational depth in Transformers. By systematically varying width–depth aspect ratios under approximately fixed parameter and training budgets, we show that the benefit of allocating more capacity to depth is strongly architecture-dependent. In particular, HC and Full AttnRes consistently benefit from deeper and narrower model shapes, while Pre-LN and most normalization-based variants show much weaker or even unfavorable depth scaling. Our layer-wise analyses further suggest that these gains are associated with more heterogeneous, consequential, and order-sensitive computation across depth, rather than simply improved trainability. We also show the hidden cost of mHC and Block AttnRes. Finally, these gains come with additional efficiency costs, highlighting an important accuracy–efficiency trade-off. Overall, our results identify residual connection design as a key determinant of whether depth can serve as a meaningful scaling axis for Transformer models.

\bibliography{example_paper}
\bibliographystyle{report}

\clearpage
\appendix

\section{Architecture-specific Settings} \label{app:architecture-settings}
All variants share the LLaMA-like backbone described in
Section~\ref{sec:architectures}. Here we record every architecture-specific choice that deviates from it.
Table~\ref{tab:arch_params} summarises the resulting parameter counts.

\paragraph{Pre-LN \cite{pre-ln}.}
The reference block $h_\ell = h_{\ell-1} + F_\ell(\mathrm{RMSNorm}(h_{\ell-1}))$
with one pre-normalisation per sub-layer and no additional parameters.

\paragraph{Sandwich-LN \cite{sandwich-ln1}.}
Each sub-layer output is normalised a second time before the residual
addition,
$h_\ell = h_{\ell-1} + \mathrm{RMSNorm}_{\text{out}}\!\big(F_\ell(\mathrm{RMSNorm}_{\text{in}}(h_{\ell-1}))\big)$,
giving four RMSNorms per block. The output norms are initialised to the
identity (unit scale). This adds $2d$ parameters per block.

\paragraph{LayerNorm Scaling (LNS) \cite{lns}.}
The normalised input of both sub-layers in block $\ell$ (1-indexed) is
multiplied by $1/\sqrt{\ell}$ before entering attention and the feed-forward
network; the same factor is used for both sub-layers of a block, so it ranges
from $1$ in the first block to $1/\sqrt{L}$ in the last (e.g., $0.20$ at $L{=}24$,
$0.18$ at $L{=}32$). LNS introduces no parameters but requires a
$5\times$ larger optimal learning rate ($1 \times 10^{-2}$) than the other variants.

\paragraph{DeepNorm \cite{deepnorm}.}
Based on Post-LN \cite{post-ln}, DeepNorm up-scales the residual connection before performing layer
normalization. 
\[
  h_\ell = \mathrm{LN}\big(\alpha\, h_{\ell-1} + F_\ell(h_{\ell-1})\big),
  \qquad \alpha = (2L)^{1/4},
\]
with $\alpha$ ranging from $2.38$ at $L{=}16$ to $2.83$ at $L{=}32$.
Following the original recipe \cite{deepnorm}, the block normalisation is a mean-centring
LayerNorm (without bias) rather than RMSNorm, while the head normalisation
remains RMSNorm.
Initialisation follows DeepNet \cite{deepnorm}: the value and output projections of attention
and all three feed-forward matrices are scaled by
$\beta = (8L)^{-1/4}$ (std $0.02, \beta \approx 0.0054$ at $L{=}24$); query,
key, embedding, and head weights keep std $0.02$.
DeepNorm is trained with optimal learning rate $1 \times 10^{-3}$.

\paragraph{KEEL \cite{keel}.}
KEEL combines pre- and post-normalisation with a large residual gain,
\[
  h_\ell = \mathrm{RMSNorm}_{\text{post}}\big(\alpha\, h_{\ell-1}
  + F_\ell(\mathrm{RMSNorm}_{\text{pre}}(h_{\ell-1}))\big),
  \qquad \alpha = \texttt{total number of sub-layers},
\]
where $\alpha$ counts attention and feed-forward sub-layers separately. 
The first block is treated specially: its attention sub-layer is a plain
Pre-LN update (no post-norm, no residual up-scaling), and its feed-forward sub-layer is
both pre-normarlised and post-normalised without residual up-scaling.
KEEL adds $d$ parameters in the first block and $2d$ in every other block.

\paragraph{AttnRes (Full) \cite{attnres}.}
Every sub-layer replaces the residual sum by a softmax-weighted mixture over
the outputs of all preceding sub-layers and the embedding, i.e.\ the sources
are individual sub-layer outputs rather than accumulated hidden states.
The attention sub-layer of block $\ell$ mixes $2\ell{+}1$ sources, the
feed-forward sub-layer $2\ell{+}2$, and the output head $2L{+}1$
($49$ at $L{=}24$).
Each sub-layer owns a learned query vector $\mathbf{w}\in\mathbb{R}^{d}$ and a
per-query RMSNorm gain applied to the keys, and scores are
$\mathbf{w}^{\top}\texttt{rmsnorm}(\mathbf{v_i})$ with softmax temperature
$1$ (no $1/\sqrt{d}$ scaling).
The mixture is RMS-normalised with the consuming sub-layer's pre-norm weights
inside the fused kernel, so the standard pre-normalisation is folded into the
mixing step, and the Language modeling (LM) head likewise consumes a mixture normalised with the LM head
norm.
All query vectors, including the LM head query, are zero-initialised, so training
starts from uniform averaging over sources.
The first block's attention reads the embedding directly and has no query.
Query vectors and key gains are subject to the global weight decay.
AttnRes adds $2d$ parameters per sub-layer plus $2d$ for the head.

\paragraph{AttnRes (Block) \cite{attnres}.}
Block AttnRes uses the same block, kernel, initialisation, and parameter
count as Full AttnRes and differs only in the depth mixing schedule.
Sub-layer outputs are accumulated into a running sum like Pre-LN. At every $b$-th
sub-layer (block size $b = (2L) / 8$: $4, 5, 6, 7, 8$ sub-layers  for $L = 16, 20, 24, 28, 32$) the
running sum is frozen as a new source and reset.
Each mixture is taken over the frozen block sums and the current running sum,
so the network is always partitioned into $2L/b = 8$ \footnote{We follow the same recipe in AttnRes report \cite{attnres} and Kimi K3 report \cite{k3} for choosing block size, i.e., fixing the number of blocks at approximately 8.} blocks and the head
mixes the embedding with eight block outputs.
For even $b$ the boundaries fall on attention sub-layers only; for odd $b$
($L{=}20$, $28$) they alternate between attention and feed-forward
sub-layers.

\paragraph{Hyper-Connections (HC) \cite{hc}.}
We use $n=4$ residual streams, with an HC connector replacing the residual
addition around each pre-normalised attention and FFN branch.
Routing combines learned static coefficients with input-dependent $\tanh$
corrections projected from each affine-free RMS-normalised stream.
Routing parameters use BF16. Dynamic projections are zero-initialised,
with two learned scalar gains initialised to $0.01$.
The initial residual map is the identity.
Zero-indexed sub-layer $k$ reads only stream $k\bmod n$ and writes its
output to all streams, with unit read and write weights.
Attention output and FFN down-projections are additionally scaled by
$1/\sqrt{n}=1/2$ at initialisation. Static routing weights receive no weight
decay, while dynamic projections use $0.1$.
Embeddings are replicated across streams, which are summed before the
final RMSNorm and LM head.

\paragraph{Manifold-Constrained Hyper-Connections (mHC) \cite{mhc}.}
\label{app:mhc-config}
We use Liger Kernel 0.8.0\footnote{\url{https://github.com/linkedin/Liger-Kernel/tree/v0.8.0}}
with $n=4$ streams and 20 Sinkhorn--Knopp iterations.
Each attention and FFN connector replaces the standard residual addition
around the pre-normalised branch. A linear projection of the affine-free
RMS-normalised, concatenated streams produces input-dependent routing
logits: read weights use sigmoid, write weights use twice sigmoid, and
residual maps use Sinkhorn normalisation, without $\tanh$.
RMS and Sinkhorn epsilons are $10^{-6}$. The read-weight epsilon is zero.
Projection weights and branch computation use BF16, while routing biases,
learned gains, and coefficient normalisation use FP32.

\begin{figure}[H]
    \centering
    \includegraphics[width=1.0\linewidth]{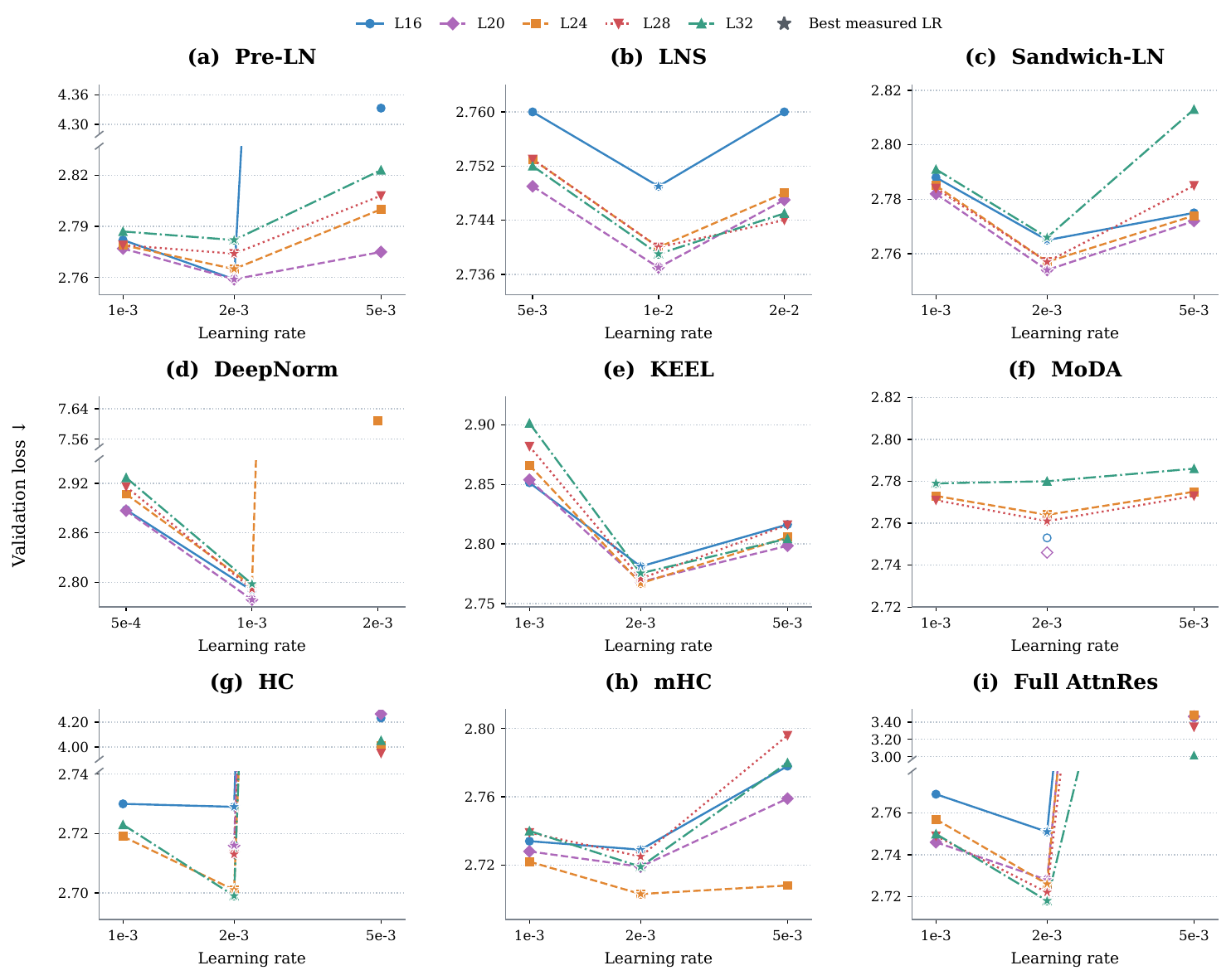}
    \caption{Learn rate sweep over the architectures. We choose the same optimal learning rate for Block AttnRes as for Full AttnRes. For mHC, we used \textsc{Legacy near-uniform init} for the learning rate sweep, which differs very slightly from the results reported in the main text using \textsc{0/-8 gap-8 init}.}
    \label{fig:lr_sweep}
\end{figure}

We initialise the dynamic projection $\Phi$ to zero and its three scalar
gains to $0.01$. At zero-indexed sub-layer $k$, read biases are $+8$ for
stream $k\bmod n$ and $-8$ otherwise; write biases are zero; residual
biases are $0$ on the diagonal and $-8$ off-diagonal (the 0/-8 GAP-8 init mentioned in \autoref{fig:lr_sweep}).
This gives approximately one-hot reads, unit writes, and near-identity
residual mixing, overriding Liger's default initialisation.
Attention output and FFN down-projection weights are scaled by
$1/\sqrt{n}=1/2$ after the base initialisation with std $0.02$.
Biases and scalar gains receive no weight decay, while $\Phi$ uses $0.1$.
Embeddings are replicated across streams, which are averaged before the
final RMSNorm and LM head.

\paragraph{Mixture-of-Depth Attention (MoDA) \cite{moda}.}
\label{app:moda-prenorm}
We use Pre-LN MoDA with the authors' \texttt{v17} Triton kernel, patched
for our GPU types and head dimensions. RMSNorm precedes both attention
and the FFN. Attention uses one softmax over causal sequence KV and
earlier-layer, same-token depth KV, with scale
$1/\sqrt{d_{\mathrm{head}}}$. Attention KV projections are reused;
FFN branches add KV projections, except in the final block where no later
layer reads them. The baseline's RoPE and additive residuals are retained.

Although the original study favours Post-LN~\citep{moda}, it optimises
poorly under our shared recipe. As shown in
Table~\ref{tab:moda-normalization}, at $L=24$ and learning rate
$2\times10^{-3}$, Post-LN MoDA ends at validation CE $3.842$ versus $2.764$
for Pre-LN MoDA. Reducing the learning rate to $1 \times 10^{-3}$ gives the best tested
Post-LN MoDA result, $2.912$, still above Pre-LN MoDA. We therefore retain Pre-LN MoDA
for the shape comparison.

\begin{table}[ht]
    \centering
    \small
    \setlength{\tabcolsep}{10pt}
    \renewcommand{\arraystretch}{1.12}
    \caption{\textbf{Validation loss for MoDA normalization variants ($L=24$ model).}}
    \label{tab:moda-normalization}
    \begin{tabular}{lccccc}
        \toprule
        Peak LR & $5\times10^{-3}$ & $2\times10^{-3}$ & $1 \times 10^{-3}$
        & $5\times10^{-4}$ & $2\times10^{-4}$ \\
        \midrule
        Pre-LN MoDA & 2.775 & \textbf{2.764} & 2.773 & 2.864 & -- \\
        Post-LN MoDA & 4.893 & 3.842 & \textbf{2.912} & 2.933 & 3.443 \\
        \bottomrule
    \end{tabular}
\end{table}

\begin{table}[ht]
  \centering
  \small
  \caption{Total parameters and overhead relative to Pre-LN at $L=24$
  ($d_{\text{model}}{=}1024$). }
  \label{tab:arch_params}
  \begin{tabular}{lrr}
    \toprule
    Architecture & Parameters & $\Delta$ vs.\ Pre-LN \\
    \midrule
    Pre-LN / LNS / DeepNorm & 405.41M & --- \\
    Sandwich-LN             & 405.46M & $+0.049$M \\
    KEEL                    & 405.45M & $+0.048$M \\
    AttnRes (Full / Block) & 405.51M & $+0.098$M \\
    HC ($n{=}4$)            & 405.70M & $+0.296$M \\
    mHC ($n{=}4$)           & 410.13M & $+4.72$M \\
    MoDA                    & 453.64M & $+48.2$M \\
    \bottomrule
  \end{tabular}
\end{table}

\section{Complete Model Configurations}\label{app:sdeep-config}

Tables~\ref{tab:aspect_ratio_configs},
\ref{tab:iso-backbone}, 
\ref{tab:depth_scaling_ladder} and \ref{tab:aspect_ratio_configs_1b}  summarize the complete model configurations used in our experiments.

For the 400M benchmark, we use $(L,d)=(24,1024)$ as the reference configuration. This is a convenient and representative Transformer shape: $d=1024$ is a standard hidden dimension, gives a head dimension of 64 with 16 attention heads, and results in a model of approximately 405M parameters under our backbone. We construct the remaining shapes around this reference by trading width for depth, extending from the shallow--wide $(16,1216)$ configuration to the substantially deeper $(70,640)$ configuration. The reference shape also serves as the anchor for our iso-backbone study, making the two experimental settings directly comparable.

For the 1.6B experiments, we choose $(L,d)=(28,2048)$ as the reference shape, matching the same Qwen3-1.7B \citep{qwen3} shape  to define our larger-scale backbone. In particular, it uses $d_{\mathrm{ff}}=6144$, 16 query heads, 8 key--value heads, and head dimension 128. We then construct two deeper and narrower variants, $(40,1728)$ and $(54,1504)$, while keeping the total model size close to 1.6B. This gives three representative aspect ratios, 73.1, 43.2, and 27.9, covering a comparable shallow-to-deep range to the main 400M study.

The multi-scale 200M--500M suite is designed to preserve approximately the same three width--depth regimes across model scales. We use aspect ratios near 28, 43, and 76, corresponding to the deep, intermediate, and shallow reference points in the 400M sweep. For example, at 400M these are exactly $(32,896)$, $(24,1024)$, and $(16,1216)$. At other parameter scales, we adjust $L$ and $d$ to reproduce these aspect ratios as closely as possible while keeping the total parameter count matched within each scale. This construction allows us to test whether the preferred width--depth allocation changes with model size rather than with a particular absolute choice of width or depth.

\input{tables/iso-backbone}
\input{tables/ladders}
\input{tables/1B_model_configs}

\section{Learning Rates \& Their Impact} \label{app:lr_sweep}

Figure~\ref{fig:lr_sweep} presents the learning-rate sweeps for the architectures evaluated in \textsc{DepthBench} across different model shapes. These results are used to select the architecture-specific learning rates adopted in our main results.

\begin{figure}[h]
    \centering
    \includegraphics[width=1.0\linewidth]{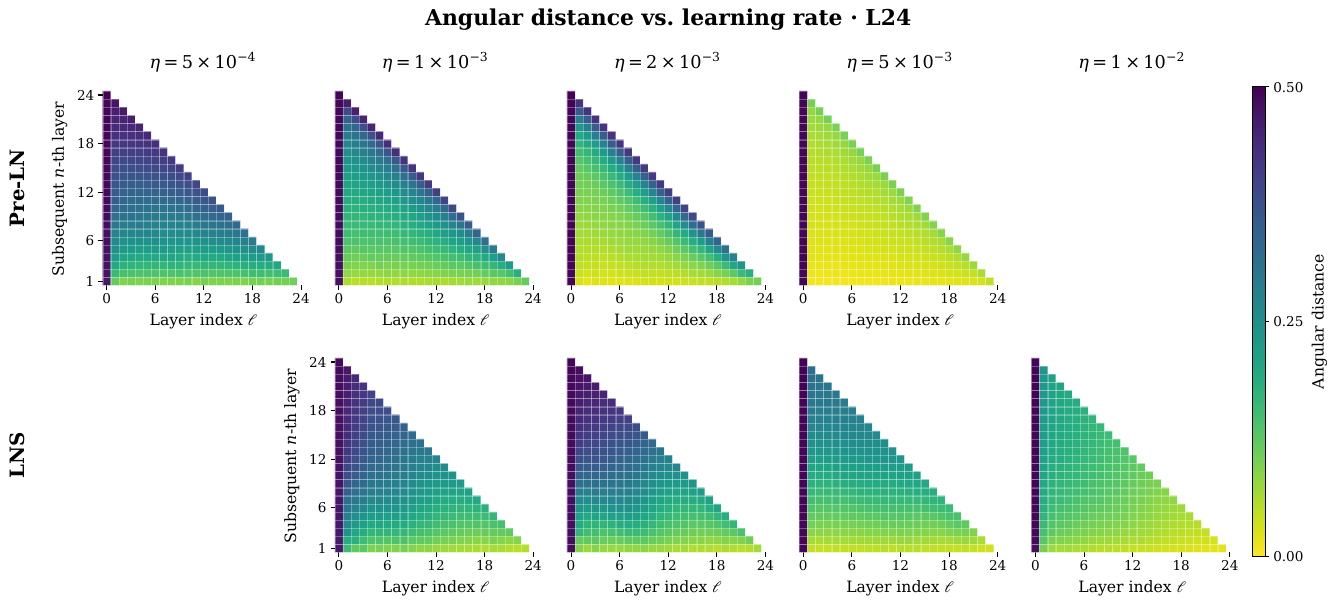}
    \caption{The analyses in Section~\ref{sec:depth-metric} can be affected by the choice of learning rate.}
    \label{fig:lr_sweep_affect}
\end{figure}

Figure~\ref{fig:lr_sweep_affect} shows the analyses in Section~\ref{sec:depth-metric} can be affected by the choice of learning rate.  While $2\times10^{-3}$ is optimal for most architectures, LNS is substantially more sensitive and achieves its best performance at a much larger learning rate of $1\times10^{-2}$, which correspondingly affects its layer-wise analysis.

\section{FLOP Accounting}
\label{app:flops}

We report forward pass FLOPs using the convention that one multiply--add counts as two FLOPs. We count the dominant matrix multiplications and vector dot products, but omit secondary operations like element-wise functions or RMSNorm. We follow the notation used throughout the paper: $T$ is the sequence length, $L=n_{\mathrm{layer}}$ is the number of Transformer layers, $d=d_{\mathrm{model}}$ is the hidden dimension, $d_{\mathrm{ff}}$ is the FFN intermediate dimension, $V$ is the vocabulary size, and $m$ is the number of residual streams for HC and mHC. We use $m=4$ in all experiments.

\paragraph{Pre-LN Transformer.} For the baseline Transformer, we count the four attention projections $Q,K,V,O$, the three SwiGLU projections, the causal attention matrix products, and the LM head. The resulting cost is
\[
F_{\mathrm{Pre\text{-}LN}}
=
L\left(8Td^2 + 6Td\,d_{\mathrm{ff}} + 2T^2d\right)
+
F_{\mathrm{head}},
\]
where
\[
F_{\mathrm{head}}
=
\begin{cases}
2TdV, & \text{for logits at all positions},\\
2dV, & \text{for final position logits}.
\end{cases}
\]
The $2T^2d$ term accounts for both $QK^\top$ and $PV$ under causal attention. Each product would cost $2T^2d$ if evaluated densely, but we follow the convention that a causal kernel evaluates approximately half of the $T^2$ query--key pairs and that the two products together cost approximately $2T^2d$. A dense implementation that evaluates the full attention matrix would instead require approximately $4T^2d$.

\paragraph{HC.} HC replaces the residual addition around each attention and FFN sublayer with a hyper-connection, giving $2L$ connectors. We count the dynamic routing projections and the read, residual mixing, and write operations, while omitting normalization and elementwise operations. The additional cost is
\[
\Delta F_{\mathrm{HC}}
=
2L\left[
2Tmd(m+1)
+
2Tmd
+
2Tm(m+1)d
+
2Tmd
\right].
\]
For $m=4$,
\[
\Delta F_{\mathrm{HC}} = 192\,TdL.
\]

\paragraph{mHC.} mHC likewise introduces one connector around each attention and FFN sub-layer. Its dominant additional cost is the projection of the flattened $md$-dimensional residual state to the $m^2+2m$ routing logits, followed by the read, $m\times m$ residual-stream mixing, and write operations. RMSNorm, sigmoid evaluations, and Sinkhorn--Knopp normalization (which is done entirely on-chip over very small matrices) are omitted under our FLOP convention. Thus,
\[
\Delta F_{\mathrm{mHC}}
=
2L\left[
2Tmd(m^2+2m)
+
2Tmd
+
2Tm^2d
+
2Tmd
\right].
\]
For $m=4$,
\[
\Delta F_{\mathrm{mHC}} = 480\,TdL.
\]
The higher FLOP cost of mHC relative to HC is therefore dominated by its $md\rightarrow m^2+2m$ routing projection rather than by the Sinkhorn iterations.

\paragraph{Full AttnRes.} Full AttnRes scores each preceding residual source using the query--source dot product $w^\top v_i$ and forms a weighted sum of the source vectors. For each source, these two operations cost $2Td$ FLOPs each, giving $4Td$ FLOPs per source. Source normalization and the softmax over depth are omitted. The first attention sub-layer reads the embedding directly and requires no depth-mixing operation. The remaining sub-layers and the LM-head readout mix $2,3,\ldots,2L+1$ sources, respectively. Hence,
\[
\Delta F_{\mathrm{Full}}
=
4Td\sum_{\ell=2}^{2L+1}\ell
=
4TdL(2L+3),
\]
and we further approximate for simplicity
\[
\Delta F_{\mathrm{Full}}
\approx
8TdL^2.
\]

\paragraph{Block AttnRes.} Block AttnRes uses the same depth-mixing operation but accumulates sub-layer outputs within blocks. Following our experimental setup, the $2L$ attention and FFN sub-layers are divided into eight blocks (more blocks are possible, but here we use 8 following their recommendations) with block size
\[
b=\frac{2L}{8}.
\]
Indexing the sub-layers by $k=0,\ldots,2L-1$, the first attention sub-layer ($k=0$) reads the embedding directly and requires no depth-mixing operation. For $k=1,\ldots,2L-1$, the mixture contains the embedding, the current running block sum, and one additional frozen source for every completed block, giving
\[
2+\left\lfloor\frac{k}{b}\right\rfloor
\]
sources. The LM head mixes the embedding with the eight completed block outputs, giving nine sources. The additional cost is therefore
\[
\Delta F_{\mathrm{Block}}
=
4Td\left[
\sum_{k=1}^{2L-1}
\left(
2+\left\lfloor\frac{k}{b}\right\rfloor
\right)
+
9
\right],
\qquad
b=\frac{2L}{8}.
\]


\section{Complete Results of Causal Score,  Permutation Score and Angular Distance} \label{heatmaps}

\newlength{\appheatw}
\setlength{\appheatw}{0.78\linewidth}

This section reports the complete layer-pair heatmaps in Section \ref{sec:depth-metric} for all 10
architectures at $L=16$, $24$, and $32$.

\textbf{Causal score}: entry $(s,\ell)$ measures the effect of skipping layer $s$ on the
update of a later layer $\ell$.

\textbf{Permutation score}: entry $(\ell_1,\ell_2)$ measures the effect of swapping
layers $\ell_1<\ell_2$; blue means the two layers are interchangeable, red that
their order matters.

\textbf{Angular distance}: entry $(\ell,n)$ is the distance between the residual state
after layer $\ell$ and the state $n$ layers later ($\ell=0$ is the embedding).

\begin{figure*}[p]
  \centering
  \includegraphics[width=\appheatw]{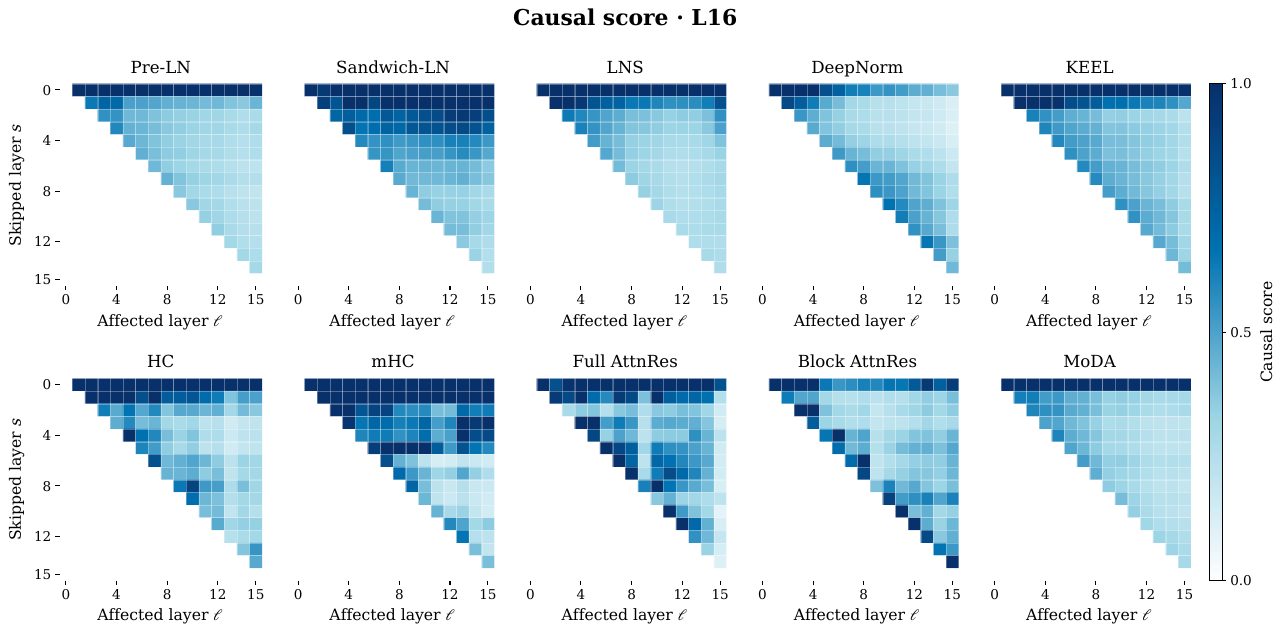}\\[6pt]
  \includegraphics[width=\appheatw]{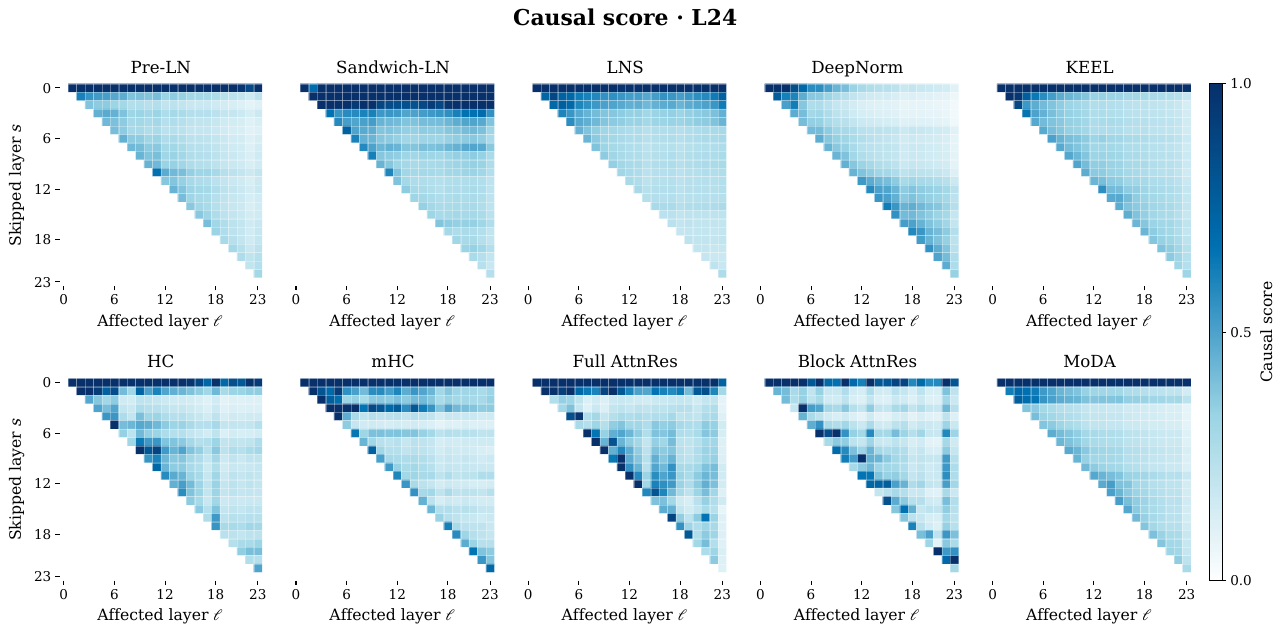}\\[6pt]
  \includegraphics[width=\appheatw]{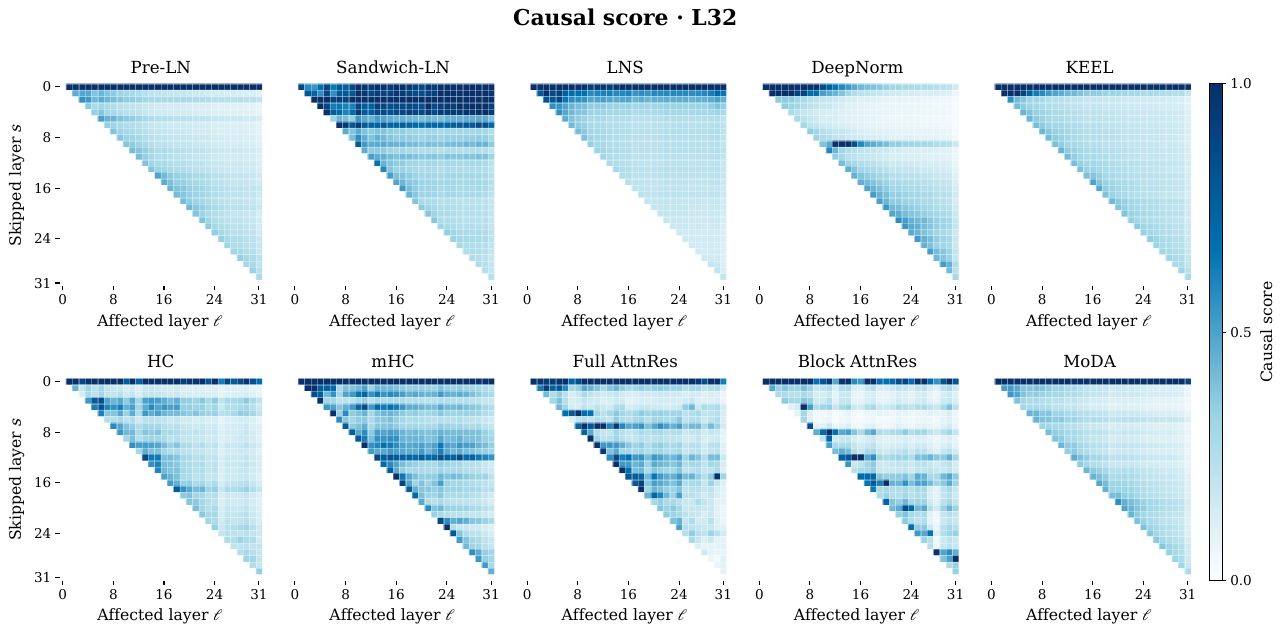}
  \caption{Causal score for all architectures at $L=16$, $24$, and $32$ (top to bottom).}
  \label{fig:app_causal}
\end{figure*}

\begin{figure*}[p]
  \centering
  \includegraphics[width=\appheatw]{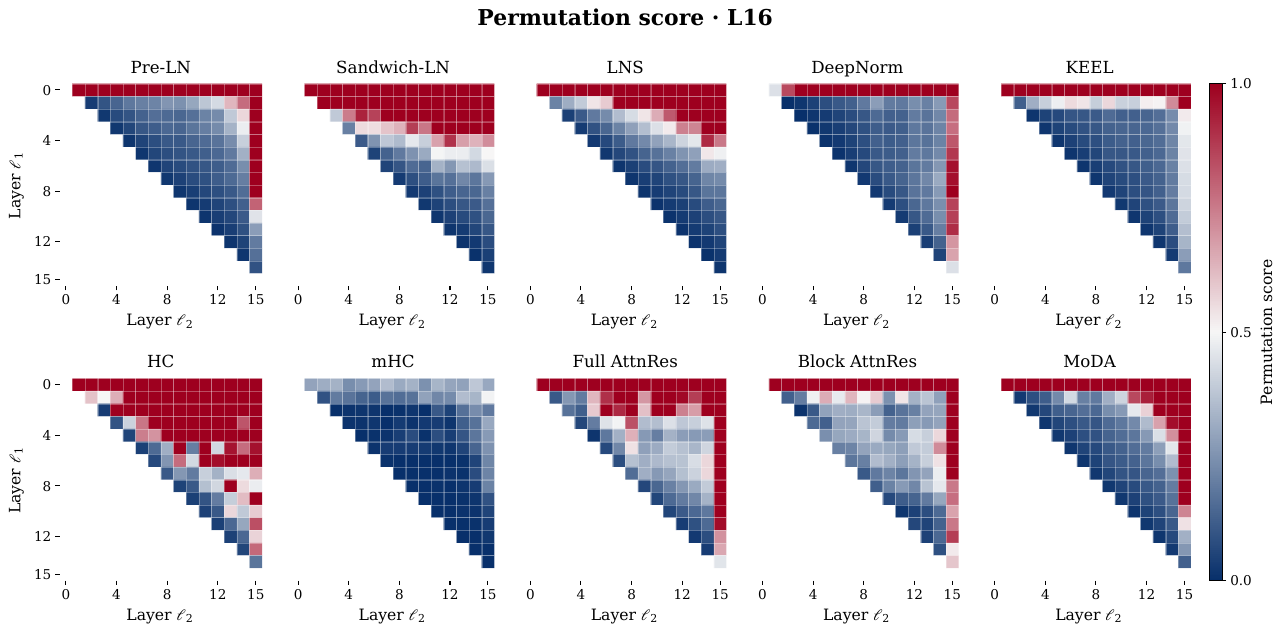}\\[6pt]
  \includegraphics[width=\appheatw]{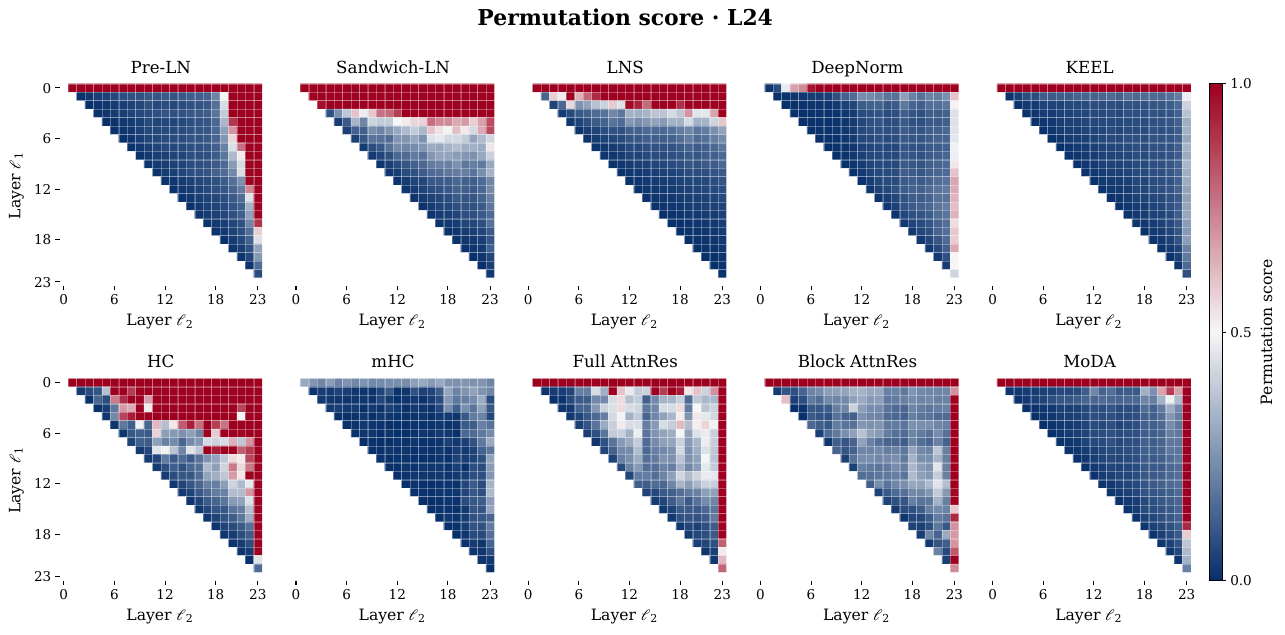}\\[6pt]
  \includegraphics[width=\appheatw]{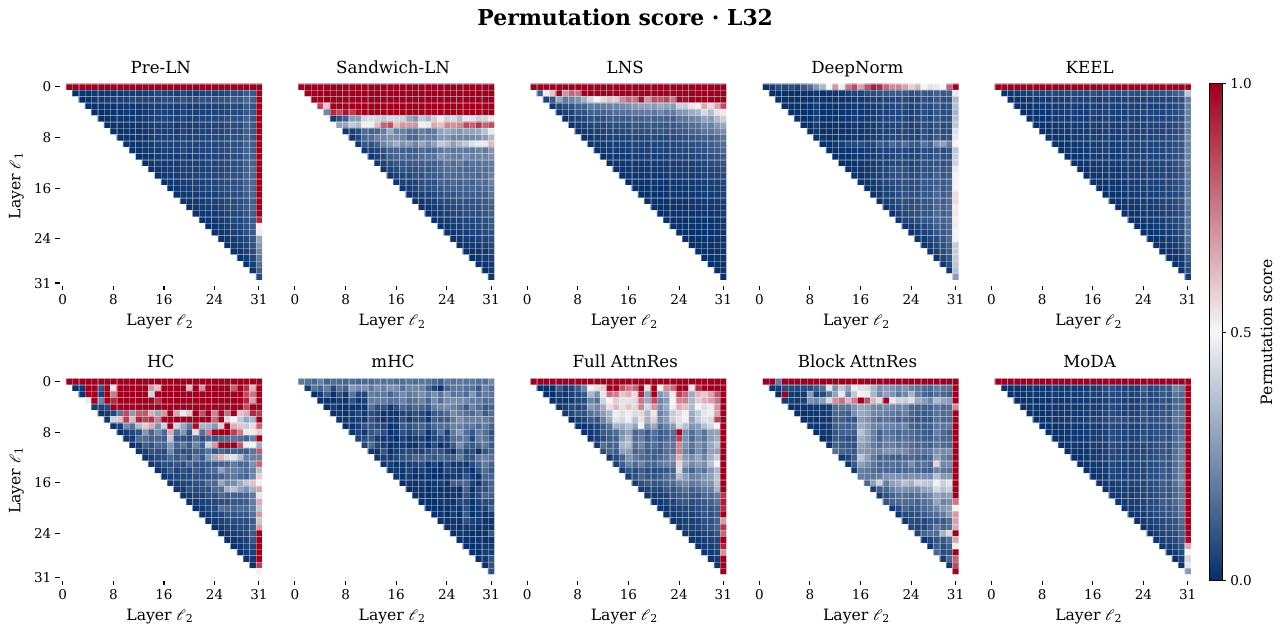}
  \caption{Permutation score for all architectures at $L=16$, $24$, and $32$ (top to bottom).}
  \label{fig:app_permutation}
\end{figure*}

\begin{figure*}[p]
  \centering
  \includegraphics[width=\appheatw]{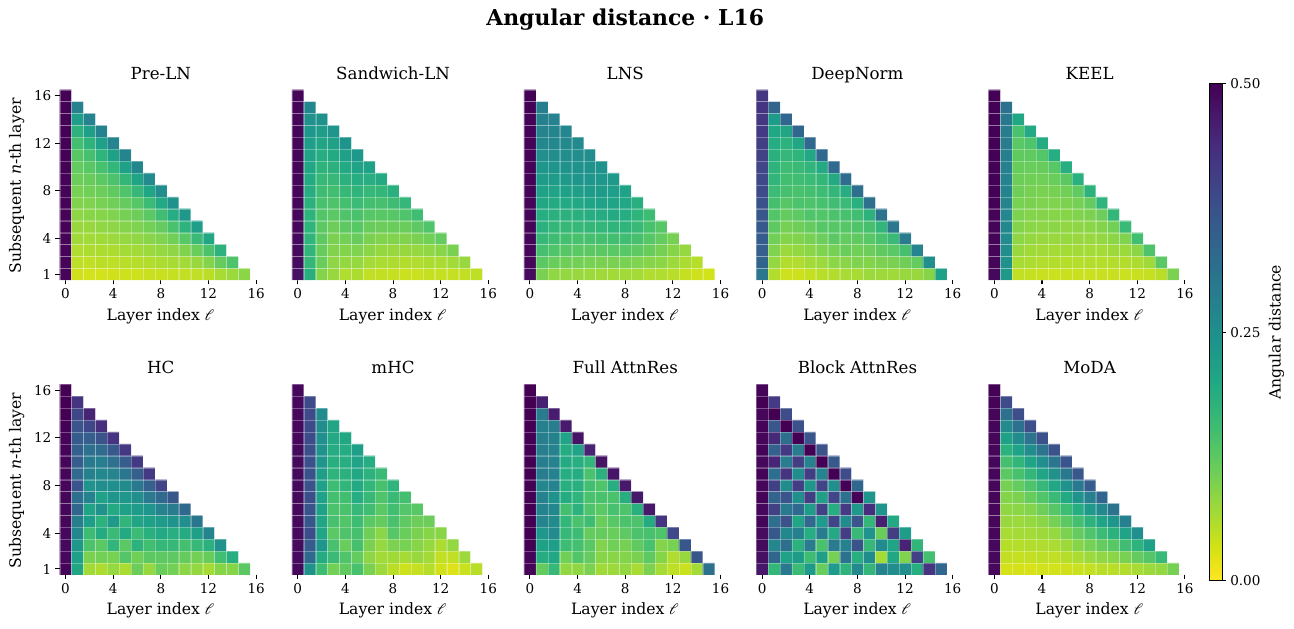}\\[6pt]
  \includegraphics[width=\appheatw]{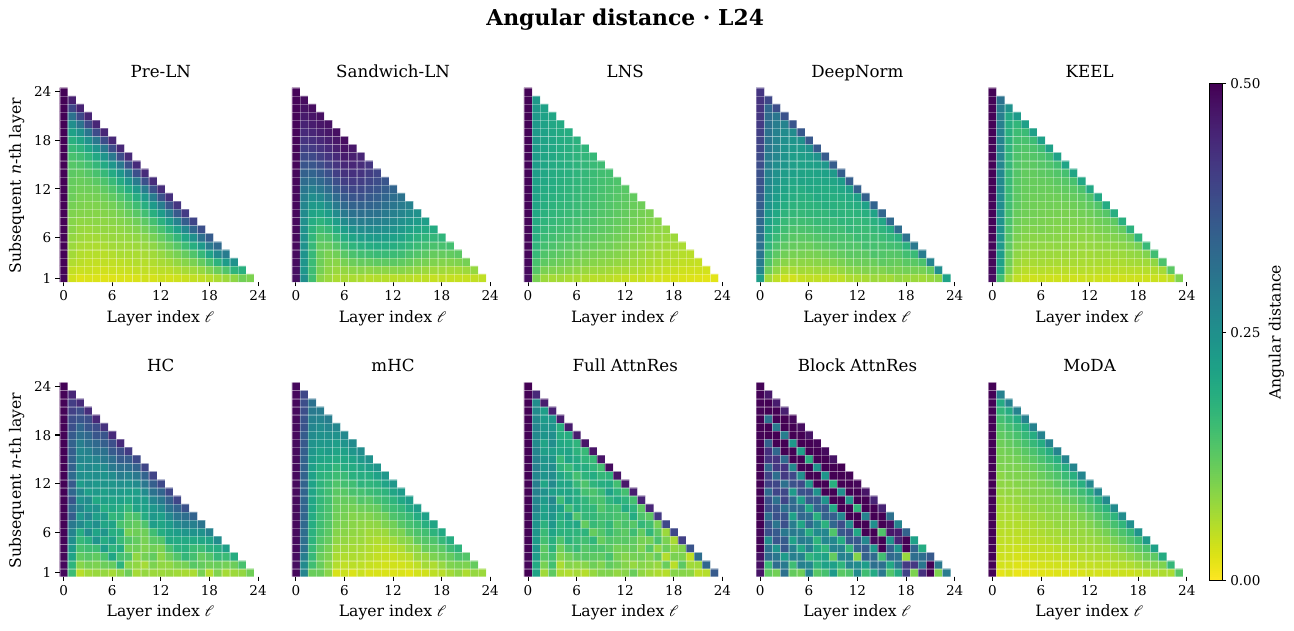}\\[6pt]
  \includegraphics[width=\appheatw]{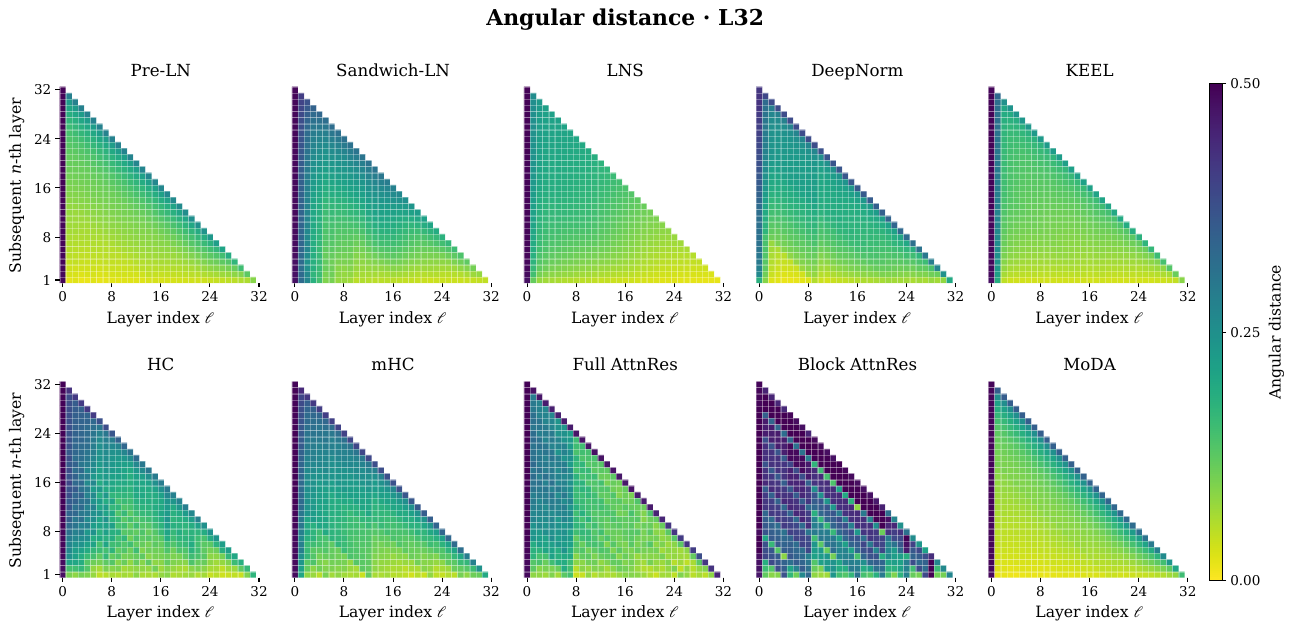}
  \caption{Angular distance for all architectures at $L=16$, $24$, and $32$ (top to bottom).}
  \label{fig:app_angular}
\end{figure*}

\end{document}

%% file: tables/architecture_sketch.tex
\begin{table}[t]
    \centering
    \small
    \setlength{\tabcolsep}{5pt}
    \renewcommand{\arraystretch}{1.18}
    \caption{
        Overview of residual update mechanisms for the architectures \citep{attnres}.
        Architectures differ primarily in how layer $\ell$ receives and aggregates
        information from preceding layers. \textit{Weight}: whether the mixing coefficients are fixed or input-dependent (dynamic). \textit{Source}: which earlier representations layer $\ell$ can access.
    }
    \label{architecture_sketch}
    \label{tab:architectures}
    \resizebox{\linewidth}{!}{
    \begin{tabular}{llll}
        \toprule
        \textbf{Method}
        & \textbf{Update rule}
        & \textbf{Weight}
        & \textbf{Source} \\
        \midrule

        \multicolumn{4}{l}{\textbf{\textit{Baseline}}} \\[-1pt]
        \hspace{1em}Pre-LN \citep{pre-ln}
        &
        $h_\ell
        =
        h_{\ell-1}
        +
        F_\ell(\mathrm{LN}(h_{\ell-1}))$
        &
        Fixed
        &
        $h_{\ell-1}$ \\

        \addlinespace[2pt]
        \midrule

        \multicolumn{4}{l}{\textbf{\textit{LayerNorm variants}}} \\[-1pt]
        \hspace{1em}Sandwich-LN \citep{sandwich-ln1}
        &
        $h_\ell
        =
        h_{\ell-1}
        +
        \mathrm{LN}_{\mathrm{out}}
        \!\left(
            F_\ell(\mathrm{LN}_{\mathrm{in}}(h_{\ell-1}))
        \right)$
        &
        Fixed
        &
        $h_{\ell-1}$ \\

        \hspace{1em}LNS \citep{lns}
        &
        $h_\ell
        =
        h_{\ell-1}
        +
        F_\ell\!\left(
            \frac{1}{\sqrt{\ell}}
            \mathrm{LN}(h_{\ell-1})
        \right)$
        &
        Fixed
        &
        $h_{\ell-1}$ \\

        \hspace{1em}DeepNorm \citep{deepnorm}
        &
        $h_\ell
        =
        \mathrm{LN}\!\left(
            \alpha h_{\ell-1}
            +
            F_\ell(h_{\ell-1})
        \right)$
        &
        Fixed
        &
        $h_{\ell-1}$ \\

        \hspace{1em}KEEL \citep{keel}
        &
        $h_\ell
        =
        \mathrm{LN}\!\left(
            \alpha h_{\ell-1}
            +
            F_\ell(\mathrm{LN}(h_{\ell-1}))
        \right)$
        &
        Fixed
        &
        $h_{\ell-1}$ \\

        \addlinespace[2pt]
        \midrule

        \multicolumn{4}{l}{\textbf{\textit{Multi-stream residuals} }} \\[-1pt]
        \hspace{1em}HC \citep{hc}
          &            
          $H_\ell             
          = H_{\ell-1}A_\ell
          + F_\ell\!\left(\mathrm{LN}(H_{\ell-1}\alpha_\ell)\right)
            \beta_\ell^\top$      
          &         
          Dynamic
          &                    
          $m$ residual streams \\
  \hspace{1em}mHC \citep{mhc}
  &
  $\begin{aligned}                          
  H_\ell                       
  &= H_{\ell-1}\mathrm{Sinkhorn}(\widetilde A_\ell)
   + F_\ell\!\left(\mathrm{LN}(H_{\ell-1}\sigma(\widetilde\alpha_\ell))\right)
     2\sigma(\widetilde\beta_\ell^\top)
  \end{aligned}$
  &                       
  Dynamic
  &                                
  $m$ residual streams \\
        \addlinespace[2pt]
        \midrule

        \multicolumn{4}{l}{\textbf{\textit{Cross-layer access} }} \\[-1pt]

        \multirow{2}{*}{\hspace{1em}AttnRes \citep{attnres}}
        &
        $\textit{Full:}\quad
        h_\ell
        \propto
        \sum_{i=0}^{\ell-1}
        \phi(w_\ell,v_i)v_i,
        \quad
        v_0 = h_1,\;
        v_{i \geq 1} = f_i(h_i)$
        &
        Dynamic
        &
        $[h_1,\ldots,h_{\ell-1}]$\vspace{0.10em}  \\

        &
        $\textit{Block:}\quad
        h_\ell
        \propto
        \sum_{i=0}^{n-1}
        \phi(w_\ell,v_i)v_i
        +
        \phi(w_\ell,v_n^{j})v_n^{j}$
        &
        Dynamic
        &
        $[v_0,\ldots,v_{n-1},v_n^{j}]$ \vspace{0.40em} \\ 


        \hspace{1em}{MoDA (Pre-LN)} \citep{moda}
          &
          $h_\ell
          = h_{\ell-1}
          + \mathrm{MoDA}_\ell\!\left(
              \mathrm{LN}(h_{\ell-1});\mathcal C_{<\ell}
            \right)$
          &
          Dynamic
          &
          $\bigl[
              h_{\ell-1,\le t};
              h_{0,t},\ldots,h_{\ell-2,t}
          \bigr]$\\

        \bottomrule
    \end{tabular}
    }
\end{table}

%% file: tables/setting_overview.tex
\begin{table}[t]
    \centering
    \small
    \setlength{\tabcolsep}{4.5pt}
    \caption{
    Overview of the experimental settings in \textsc{DepthBench}.
    }
    \label{tab:experiment_overview}
    \resizebox{\linewidth}{!}{
    \begin{tabular}{@{}lcccl@{}}
        \toprule
        \textbf{Setting}
        & \textbf{Matched Budget}
        & \textbf{\# Shapes}
        & \textbf{Aspect Ratios}
          & \textbf{Purpose} \\
        \midrule

        Main benchmark
        & $\approx$400M total
        & 7
        & $9.1$--$76.0$
        & Comprehensive architecture comparison \\

        Iso-backbone control
        & $\approx$300M backbone
        & 6
        & $6.5$--$78.0$
        & Control for backbone-size variation \\

        Multi-scale suite
        & 200M--500M total
        & 3 
        & $\approx$28, $\approx$43, $\approx$76
        & Test consistency across model scales \\

        Validation at scale
        & $\approx$1.6B total
        & 3
        & $27.9,\ 43.2,\ 73.1$
        & Validate trends at larger scale \\

        \bottomrule
    \end{tabular}
    }
\end{table}

%% file: tables/main_model_configs.tex
\begin{table}[t]
\centering
\caption{400M model configurations across different width--depth aspect ratios.}
\label{tab:aspect_ratio_configs}
\small
\setlength{\tabcolsep}{4pt}  
\begin{tabular*}{\linewidth}{@{\extracolsep{\fill}}ccccccccc@{}}
\toprule
Layers & Hidden & Intermediate & Heads & Head Dim. & Aspect Ratio & Backbone Size & Total Size & Total Diff. \\
\midrule
16 & 1216 & 3248 & 16 & 76 & 76.00 & 284M & 407M & $+0.28\%$ \\
20 & 1120 & 2992 & 16 & 70 & 56.00 & 301M & 414M & $+2.14\%$ \\
24 & 1024 & 2736 & 16 & 64 & 42.67 & 302M & 405M & $0.00\%$ \\
28 & 960  & 2560 & 16 & 60 & 34.29 & 310M & 406M & $+0.21\%$ \\
32 & 896  & 2400 & 16 & 56 & 28.00 & 309M & 399M & $-1.49\%$ \\
42 &  800 & 2144 & 16 & 50 & 19.05 & 324M & 404M & $-0.31\%$ \\
70 &  640 & 1712 & 16 & 40 &  9.14 & 345M & 409M & $+0.94\%$ \\
\bottomrule
\end{tabular*}
\end{table}

%% file: tables/optimal_learning_rate.tex
\begin{table}[t]
\centering
\caption{Optimal learning rate across architectures.}
\label{tab:optimal_lr}
\small
\resizebox{\textwidth}{!}{
\begin{tabular}{*{10}{c}}
\toprule
 Pre-LN & Sandwich-LN & LNS & DeepNorm & KEEL & MoDA & Full AttnRes & Block AttnRes & HC & mHC  \\
\midrule
 $2 \times 10^{-3}$ & $2 \times 10^{-3}$ & $1 \times 10^{-2}$ & $1 \times 10^{-3}$ & $2 \times 10^{-3}$ & $2 \times 10^{-3}$ & $2 \times 10^{-3}$ & $2 \times 10^{-3}$ & $2 \times 10^{-3}$ & $2 \times 10^{-3}$ \\
\bottomrule
\end{tabular}
}
\end{table}

%% file: tables/iso-backbone.tex
\begin{table}[h]
    \centering
    \caption{Iso-backbone 300M model configurations}
    \label{tab:iso-backbone}
    \small
        \setlength{\tabcolsep}{2.5pt}

    \begin{tabular}{ccccccccc}
        \toprule
        Layers & Hidden & Intermediate & Heads & Head Dim. &
        Aspect Ratio & Backbone Size & Backbone Diff. & Total Size \\
        \midrule
        16 & 1248 & 3328 & 16 & 78 & 78.00 & 299M & $-1.11\%$ & 425M \\
        24 & 1024 & 2736 & 16 & 64 & 42.67 & 302M & $\phantom{+}0.00\%$ & 405M \\
        32 &  896 & 2400 & 16 & 56 & 28.00 & 309M & $+2.26\%$ & 399M \\
        42 &  768 & 2048 & 16 & 48 & 18.29 & 297M & $-1.69\%$ & 375M \\
        70 &  608 & 1632 & 16 & 38 &  8.69 & 312M & $+3.15\%$ & 373M \\
        84 &  544 & 1456 & 16 & 34 &  6.48 & 299M & $-0.99\%$ & 354M \\
        \bottomrule
    \end{tabular}
\end{table}

%% file: tables/ladders.tex
\begin{table}[h]
\centering
\caption{Model configurations for the depth scaling ladder across 200M--500M scales.}
\label{tab:depth_scaling_ladder}

\small
\setlength{\tabcolsep}{2.5pt}

\begin{tabular*}{\linewidth}{@{\extracolsep{\fill}}ccccccccc@{}}
\toprule
Layers & Hidden & Intermediate & Heads & Head Dim. &
Aspect Ratio & Backbone Size & Total Size & Total Diff. \\
\midrule

\multicolumn{9}{l}{\textbf{200M -- 4B training tokens}} \\
24 & 672  & 1792 & 16 & 42 & 28.00 & 130M & 198M & $-3.42\%$ \\
18 & 768  & 2048 & 16 & 48 & 42.67 & 127M & 205M & $0.00\%$ \\
12 & 896  & 2400 & 16 & 56 & 74.67 & 116M & 206M & $+0.69\%$ \\
\midrule
\addlinespace[3pt]
\multicolumn{9}{l}{\textbf{300M -- 6B training tokens}} \\
28 & 800  & 2144 & 16 & 50 & 28.57 & 216M & 296M & $+1.09\%$ \\
21 & 896  & 2400 & 16 & 56 & 42.67 & 203M & 293M & $0.00\%$ \\
14 & 1056 & 2816 & 16 & 66 & 75.43 & 187M & 294M & $+0.18\%$ \\
\midrule
\addlinespace[3pt]
\multicolumn{9}{l}{\textbf{400M -- 8B training tokens}} \\
32 & 896  & 2400 & 16 & 56 & 28.00 & 309M & 399M & $-1.49\%$ \\
24 & 1024 & 2736 & 16 & 64 & 42.67 & 302M & 405M & $0.00\%$ \\
16 & 1216 & 3248 & 16 & 76 & 76.00 & 284M & 406M & $+0.28\%$ \\
\midrule
\addlinespace[3pt]
\multicolumn{9}{l}{\textbf{500M -- 10B training tokens}} \\
34 & 992  & 2656 & 16 & 62 & 29.18 & 403M & 502M & $-0.42\%$ \\
26 & 1120 & 2992 & 16 & 70 & 43.08 & 392M & 504M & $0.00\%$ \\
17 & 1344 & 3584 & 16 & 84 & 79.06 & 368M & 504M & $-0.16\%$ \\

\bottomrule
\end{tabular*}

\end{table}

%% file: tables/1B_model_configs.tex
\begin{table}[h]
\centering
\caption{1.6B model configurations across different width--depth aspect ratios.}
\label{tab:aspect_ratio_configs_1b}

\small
    \setlength{\tabcolsep}{2.5pt}

\begin{tabular*}{\linewidth}{@{\extracolsep{\fill}}ccccccccc@{}}
\toprule
Layers & Hidden & Intermediate & Q/KV Heads & Head Dim. &
Aspect Ratio & Backbone Size & Total Size & Total Diff. \\
\midrule
28 & 2048 & 6144 & 16 / 8 & 128 & 73.14 & 1.409B & 1.615B & $0.00\%$ \\
40 & 1728 & 5184 & 16 / 8 & 108 & 43.20 & 1.433B & 1.607B & $-0.51$\% \\
54 & 1504 & 4512 & 16 / 8 & 94  & 27.85 & 1.466B & 1.617B & $+0.11\%$ \\
\bottomrule
\end{tabular*}

\end{table}